\documentclass[10pt,leqno]{amsart}
\usepackage{graphicx}
\usepackage[utf8]{inputenc}
\usepackage[T1]{fontenc}
\usepackage{indentfirst,csquotes}
\usepackage{newunicodechar}
\usepackage{multirow}

\usepackage{amssymb,amsthm,amsmath}
\usepackage{xcolor,paralist,tcolorbox,booktabs,array,tabularx,subfig,fancyhdr,etoolbox}
\usepackage{hyperref}
\newcommand{\missingfigure}[1]{%
  \fbox{\parbox[c][0.22\textheight][c]{0.9\linewidth}{%
    \centering Missing figure file:\\[0.5ex]\texttt{\detokenize{#1}}}}%
}

\makeatletter
\providecommand{\@secnumpunct}{.}
\renewcommand\section{\@startsection{section}{1}{\z@}%
  {2ex plus .5ex minus .2ex}{1ex plus .2ex}{\normalfont\Large\bfseries}}
\renewcommand\subsection{\@startsection{subsection}{2}{\z@}%
  {1.5ex plus .4ex minus .2ex}{.8ex plus .2ex}{\normalfont\large\bfseries}}
\makeatother

\hypersetup{ colorlinks=true, linkcolor=black, filecolor=black, urlcolor=black }

\usepackage{lipsum}

\begin{document}

\title[HiPerViT: A Hierarchical Perceiver--Vision Transformer]
{HiPerViT: A Hierarchical Perceiver--Vision Transformer Architecture 
for Multi-Scale Texture Recognition}

\maketitle

\vspace{-2.5em}

\begin{center}

{\large
Jo{\~a}o Pedro C. A. de S{\'a}$^{1}$
\quad and \quad
Odemir Martinez Bruno$^{1,2}$
}

\vspace{0.8em}

{\small
$^{1}$Institute of Mathematics and Computer Sciences (ICMC),
University of S{\~a}o Paulo (USP),\\
Avenida Trabalhador S{\~a}o-carlense, 400,
S{\~a}o Carlos, SP 13566-590, Brazil

\vspace{0.6em}

$^{2}$S{\~a}o Carlos Institute of Physics (IFSC),
University of S{\~a}o Paulo (USP),\\
Avenida Trabalhador S{\~a}o-carlense, 400,
S{\~a}o Carlos, SP 13566-590, Brazil

\vspace{0.8em}

\texttt{joaosa@usp.br} (J. P. C. A. de S{\'a})\\
\texttt{bruno@ifsc.usp.br} (O. M. Bruno)
}

\end{center}

\vspace{1em}


\begin{abstract}
Texture recognition remains challenging for modern vision models because discriminative evidence is often carried by higher-order spatial statistics rather than by object shape alone. While Vision Transformers provide strong long-range modeling capacity, their standard object-centric representations do not explicitly expose such statistical structure, which limits texture sensitivity in fine-grained recognition settings. We present \textbf{HiPerViT}, a compact vision-only architecture that injects an explicit second-order statistical prior into a transformer-based recognition pipeline. The method combines global and local image views with a compact bilinear descriptor encoded as a \emph{statistical token}, and integrates this token with first-order spatial representations through Perceiver-style latent distillation. This design enables direct interaction between spatial tokens and second-order feature co-occurrence statistics, providing the model with explicit access to texture-relevant information without requiring multimodal pretraining or ensemble construction. Across six texture recognition benchmarks, HiPerViT achieves consistent improvements over strong vision-only baselines under the reported evaluation protocols, including gains of +3.05 percentage points on DTD, +10.48 on GTOS-Mobile, and +10.10 on 1200Tex. Beyond benchmark performance, our analyses show that these gains are largely invariant to the backbone depth used to extract second-order statistics and to the ordering of interaction and distillation stages. This pattern suggests that the primary source of improvement is not a specific fusion topology, but the explicit availability of second-order statistical information as a first-class representational signal. These results support explicit statistical tokenization as an effective and robust design principle for texture-centric visual recognition. 
\end{abstract} 

\bigskip

\section{Introduction}

Texture recognition remains a tough problem for modern vision systems because evidence is often encoded in higher-order spatial statistics rather than object shape alone. In medical imaging, microscopy, remote sensing, robotics, and industrial inspection, successful recognition depends less on canonical object form than on repeated micro-structures, stochastic organization, and feature co-occurrence patterns distributed across space \cite{Haralick1973,Cimpoi2015}. This makes texture recognition different from standard object-centric categorization: robust performance requires sensitivity to local structure, long-range context, and statistical organization across scales. 

This distinction exposes a limitation of current visual architectures. Convolutional neural networks (CNNs) provide strong locality and translation priors but don't preserve higher-order statistical structure as a first-class representation. Vision Transformers (ViTs) offer a mechanism for modeling long-range interactions through self-attention, yet their standard training regime encourages representations that are effective for global structural recognition without exposing co-occurrence statistics \cite{Dosovitskiy2021,Geirhos2019}. Large-scale multimodal pre-training can broaden the scope of learned representations, but such strategies are expensive and not always practical in specialized settings with limited labeled data and domain-specific supervision. 

A central question remains unresolved: when transformer-based models improve on texture recognition, do gains arise from a particular fusion architecture or from restoring explicit access to statistical information? Much of the recent literature combines multiscale processing, token mixing, pooling, or auxiliary branches, but this often leaves unclear which ingredient is responsible for the improvement. From a representation design standpoint, the issue is whether texture-sensitive recognition benefits primarily from a specific architectural composition or from exposing higher-order statistics as an explicit representational signal. 

In this work, we address this question through \textbf{HiPerViT}, a vision-only architecture that augments transformer representations with an explicit second-order statistical prior. HiPerViT combines global and local image views, encodes compact bilinear feature co-occurrence statistics as a statistical token \cite{Gao2016}, and integrates this token with first-order spatial representations through a Perceiver-style latent bottleneck \cite{Jaegle2021}. The resulting design enables direct interaction between spatial tokens and second-order statistical information within a unified fusion pipeline. Importantly, the goal is not to replace first-order visual structure but to make higher-order statistical evidence explicitly available when it is informative for the task. 

This representation-centered framing motivates a testable hypothesis. If HiPerViT improves mainly because it uses a particularly favorable fusion design, then its performance should depend strongly on where interaction and latent distillation are placed in the pipeline. If, instead, the main benefit comes from making second-order statistical information explicitly available to the model, then the gain should remain relatively stable across reasonable changes in fusion ordering and in the backbone depth used to extract the statistical descriptor. If observed gains were mainly due to a privileged fusion topology, performance should be highly sensitive to interaction ordering and latent distillation. If they were mainly due to extracting second-order descriptors at a favorable stage of the backbone, performance should vary strongly with extraction depth. By contrast, if the principal benefit comes from exposing explicit second-order information as a first-class representational signal, then improvements should be robust to both fusion ordering and extraction depth. Our experiments distinguish among these alternatives. 

Across six texture recognition benchmarks, HiPerViT achieves competitive or best-reported top-1 accuracy while remaining computationally compact. Under the reported protocols, it reaches 93.05\% on DTD, 94.58\% on GTOS-Mobile, and 97.50\% on 1200Tex, and in matched-protocol comparisons it matches or exceeds transformer baselines such as ViT-B/16, DeiT, Swin, and VORTEX on all six datasets. Empirical results support the representation-level interpretation above. The gains associated with the statistical token are stable across alternative fusion topologies and backbone depths used to extract second-order statistics.  We show that the contribution of the statistical token is strongest in low-data regimes, consistent with the view that explicit second-order tokenization acts as a conditional inductive bias. Cue-conflict analyses indicate that adding the statistical token does not overturn the dominant object-centric perceptual preference of the underlying backbone, suggesting that the proposed mechanism expands texture sensitivity without destabilizing the primary representational prior. 

Taken together, the results support a representation-level conclusion. The low-data experiments show that the statistical token is most helpful when supervision is limited, the topology ablations show that the gain persists across different interaction--distillation orderings, and the capture-depth analyses show that final accuracy remains stable even when the raw second-order descriptors differ markedly in standalone separability. Together, these findings suggest that HiPerViT does not depend on a single privileged fusion layout; rather, its main advantage comes from making second-order statistical structure explicitly available for interaction with first-order spatial tokens. HiPerViT provides one efficient realization of this idea, but the broader implication is methodological: explicit statistical tokenization is a robust design principle for texture-oriented visual recognition.

Our main contributions can be summarized as follows: (i) we introduce HiPerViT, a compact vision-only architecture that integrates an explicit second-order statistical token through multiscale token fusion and Perceiver-style latent distillation; (ii) we show that HiPerViT achieves consistently strong performance across six texture recognition benchmarks while remaining more compact than larger vision-only comparison models; (iii) we demonstrate that the contribution of the statistical token is robust to both the backbone depth at which second-order statistics are extracted and the choice of fusion topology, indicating that the improvement arises primarily from the availability of explicit second-order information rather than from a particular architectural configuration; and (iv) we show that the statistical token acts as a conditional inductive bias, providing its strongest contribution under limited supervision while leaving dominant first-order perceptual priors largely intact in object-centric settings.

\section{Background}

The challenge of texture recognition has long fascinated both neuroscientists and computer vision researchers. The human visual system can effortlessly distinguish velvet from denim or granite from sand, even under dramatic changes of scale or illumination. Understanding how to replicate this capability computationally has led to successive paradigm shifts, each echoing, in different ways, the hierarchical computations of biological vision.

\subsection{Statistical Descriptors of Texture}
The first wave of computational approaches sought to capture texture through explicit statistics, inspired by the idea that perception relies on the distribution of simple local features. Gabor filters, for example, model orientation- and frequency-selective responses reminiscent of V1 receptive fields \cite{Hubel1962, Fogel1989}. Gray-Level Co-occurrence Matrices (GLCM) \cite{Haralick1973} quantified how pixel intensities co-occur, encoding regularities that correlate with human judgments of roughness, smoothness, or directionality. Local Binary Patterns (LBP) later distilled these ideas into compact codes of local contrast.

These descriptors demonstrated that even simple statistics could reproduce aspects of texture perception and perform reliably on controlled datasets. Yet, their handcrafted nature limited adaptability: descriptors tuned to specific frequency ranges or co-occurrence patterns often failed to generalize across scales, lighting conditions, or domains.

A breakthrough emerged when researchers extended this statistical view beyond first-order and marginal distributions to explicitly model pairwise interactions. The bilinear pool represented each image as the outer product of the feature maps, capturing second-order correlations between all channels in a translation-invariant way \cite{Lin2015}. This paralleled findings in neuroscience that mid-level cortical areas (V2, V4) integrate co-occurrence patterns of oriented filters to represent texture surfaces \cite{Freeman2013}. However, the full bilinear representation was prohibitively large (e.g., 262k dimensions for 512 channels). Compact Bilinear Pooling (CBP) \cite{Gao2016} addressed this by sketching the outer product into a few thousand dimensions with minimal accuracy loss, enabling efficient training and deployment. Refinements such as iSQRT-COV \cite{Li2018} and Shifted Random Maclaurin pooling \cite{Yu2021} further optimized this principle, consolidating the role of second-order statistics as a cornerstone of texture recognition.

In summary, from early co-occurrence matrices to modern compact bilinear pooling, the statistical tradition established a durable insight: textures are not merely distributions of local features, but structured arrangements whose identity often lies in the correlations among them. This statistical perspective paved the way for the next generation of approaches, where hierarchical feature learning could be naturally combined with statistical modeling.

\subsection{CNN Architecture for Texture Analysis}
The advent of deep learning radically shifted the field. CNNs \cite{LeCun1998} offered hierarchical representations learned directly from data, mirroring the layered architecture of the visual cortex. Convolutional layers acted as learned filter banks, while deeper layers captured increasingly abstract structures. Researchers quickly realized that these features, although trained for object recognition, could be repurposed for material and texture classification.

Seminal works such as FV-CNN \cite{Cimpoi2015} treated convolutional responses as dense filter banks and aggregated them orderlessly, achieving invariance to position and scale. Bilinear CNNs \cite{Lin2015} went further by encoding all pairwise channel interactions, significantly improving fine-grained recognition. Multiscale extensions (e.g., MOP-CNN, MS-CNN) highlighted another principle from biology: robust texture understanding requires integrating both micro- and macro-patterns across scales.

Despite these successes, CNN-based solutions often demanded task-specific modifications, dilated kernels for high-frequency textures, spectral branches for stationary patterns, revealing that their inductive biases were not inherently tuned for texture statistics.

\subsection{From Sequence Transformers to Vision Transformers}
The Transformer architecture originated in natural language processing with Attention Is All You Need \cite{Vaswani2017}, which replaced recurrence with self-attention to model long-range dependencies in sequential data. The key innovation was the multi-head self-attention mechanism, allowing each token to attend to every other token through learned query–key–value interactions. This design provided parallelism, scalability, and global context modeling, advantages that soon motivated exploration beyond language.

Early visual adaptations sought to translate these principles to images by tokenizing spatial patches rather than words. Dosovitskiy et al. (2021) \cite{Dosovitskiy2021} introduced the Vision Transformer (ViT), showing that with sufficient data and computation, a pure transformer could rival convolutional networks for image classification. ViT dispensed with spatial inductive biases such as locality and translation equivariance, instead relying on large-scale pre-training to learn them implicitly.

Subsequent architectures progressively reintroduced structure and efficiency: DeiT \cite{Touvron2021} improved data efficiency through distillation; Swin Transformer \cite{Liu2021} implemented hierarchical windowed attention to recover multi-scale locality; and Perceiver \cite{Jaegle2021} extended the transformer to high-dimensional or multimodal inputs via a latent bottleneck with asymmetric cross-attention, drastically reducing the cost of global attention.

This lineage, from sequence models to ViT and hierarchical or latent variants, marks a clear evolution: from global attention without structure toward architectures that integrate scale, hierarchy, and efficiency. HiPerViT continues this trajectory by uniting a multi-scale ViT backbone with a Perceiver-style latent bottleneck and compact bilinear pooling, forming a biologically inspired and computationally efficient model for texture understanding.

\subsection{The Transformer Era in Texture Recognition}
With the success of ViTs in large-scale image classification, researchers soon began investigating their suitability for texture and material recognition, domains traditionally dominated by convolutional networks and handcrafted statistical models. The hope was that the global self-attention of transformers could unify local micro-patterns and long-range contextual cues in a single framework.

However, empirical analyses revealed that vanilla ViTs develop strong shape priors, often favoring object contours over fine-grained texture cues~\cite{Geirhos2022, Naseer2021}. This effect became particularly evident on fine-grained material datasets.

To address this, several transformer variants introduced inductive biases that restore local sensitivity. VORTEX \cite{Scabini2025} extracted multiscale tokens from frozen ViTs and applied orderless second-order pooling to capture high-frequency texture cues. DBTrans \cite{Liu2024} fused global and local branches to balance contour awareness with texture detail. Other approaches combined transformers with convolutional stems or spectral encoders to re-introduce spatial priors lost in pure self-attention.

Parallel to these efforts, multimodal pre-training frameworks such as CLIP \cite{Radford2021}
 demonstrated that coupling images with textual supervision could improve robustness and partially recover texture sensitivity. Yet these gains came at the cost of billions of parameters, heavy compute budgets, and dependence on language data, conditions rarely feasible in specialized domains like medical imaging or remote sensing.

 This evolution exposed a fundamental trade-off: transformers offer unparalleled global reasoning but tend to neglect the local statistical structure that defines textures. Bridging this divide requires architectures that preserve hierarchical multiscale features while maintaining transformer-level global context.

HiPerViT addresses this gap by integrating a multi-scale ViT backbone with a Perceiver-style latent bottleneck and compact bilinear pooling, enabling explicit second-order statistical modeling within a transformer framework. Sec.~3 discusses the biological intuitions that guided this design.

\section{Biological Motivation}
Human and non-human primates have evolved a visual system that excels at distinguishing textures under variations of scale, illumination, and context, precisely the challenges that computational models still struggle to overcome. Classical neurophysiology revealed that the early visual cortex (V1) acts as a filter bank, decomposing the retinal input into localized, orientation and frequency selective channels, not unlike Gabor filters \cite{Hubel1962}. This stage emphasizes micro-patterns: edges, gratings, and fine textures. However, vision does not operate as a strictly linear pipeline. Signals are processed in parallel, distributed across specialized cortical streams \cite{Zeki1993}, ensuring that multiple scales and modalities of information can be extracted simultaneously.

As inputs propagate to mid-level areas such as V2 and V4, neurons integrate co-occurrence statistics, curvatures, and surface properties, enabling representations that capture both the fine statistics of texture and its global layout \cite{Riesenhuber1999}. Evidence indicates that these areas act as hubs of multiscale integration, combining local filter responses with broader contextual cues \cite{Freeman2013}. Such processing reflects not only hierarchical depth but also parallel specialization: different cortical regions emphasize distinct visual attributes (e.g., color, form, motion), a principle known as functional specialization \cite{Zeki1993}. In this framework, texture perception emerges from the coordinated interaction of specialized modules rather than from a monolithic sequence of transformations.

At higher cortical stages, notably the inferotemporal (IT) cortex, representations become increasingly abstract and tolerant, merging local structure with semantic associations. Importantly, the integration of parallel streams into compact codes ensures both robustness and efficiency: vast amounts of sensory data are condensed into latent representations that remain discriminative under clutter, occlusion, and variability.

These insights guided the design of HiPerViT. Its multi-scale ViT backbone mirrors the hierarchical and parallel nature of cortical processing. The Compact Bilinear Pooling module is inspired by pairwise co-occurrence coding in mid-level vision \cite{Portilla2000}. The Perceiver-style latent bottleneck is analogous to the convergent compression observed in IT cortex, condensing distributed multi-stream inputs into compact, discriminative codes.

\section{Model Architecture and Methodology}

Our central hypothesis is that robust visual recognition in micro-structure-sensitive domains requires injecting second-order statistical structure directly into the transformer's representational space. This is operationalized through \textbf{Statistical Token Injection (STI)}, defined below. The remaining architectural components, multi-scale extraction, Perceiver-style distillation, and compact bilinear pooling, serve as the supporting infrastructure that makes STI practical and scalable.

\subsection{Statistical Token Injection (STI)}

Statistical Token Injection (STI) is a representation mechanism that converts compact second-order feature statistics into an explicit token that can participate directly in transformer self-attention. Given a feature map $\mathbf{F} \in \mathbb{R}^{N \times D}$ extracted from a backbone layer, STI first computes a compact second-order descriptor $\mathbf{z}_{\mathrm{srm}} \in \mathbb{R}^{M}$ using a Count Sketch bilinear projection, where $M \ll D^2$. This operation provides a computationally efficient approximation of pairwise feature co-occurrence statistics without explicitly constructing the full outer-product representation.

The resulting descriptor is projected into the same embedding space as the spatial tokens:

\begin{equation}
\mathbf{T}_{\mathrm{srm}} = W_e \mathbf{z}_{\mathrm{srm}} + \mathbf{b}_e, \qquad \mathbf{T}_{\mathrm{srm}} \in \mathbb{R}^{D}, \end{equation}

where $W_e \in \mathbb{R}^{D \times M}$ and $\mathbf{b}_e \in \mathbb{R}^{D}$ are learnable parameters. The projected representation $\mathbf{T}_{\mathrm{srm}}$ is then inserted as a first-class token into the spatial token sequence. Consequently, standard multi-head self-attention can jointly operate on first-order spatial representations and explicit second-order statistical information.

This formulation differs from conventional bilinear heads or late-fusion strategies, in which second-order descriptors are typically appended only after spatial representation learning has already occurred. In STI, the statistical representation becomes part of the attention process itself. Spatial tokens can therefore attend to global feature co-occurrence statistics, while the statistical token can be contextually modulated by the spatial structures present in the image. We refer to this bidirectional exchange between first- and second-order representations as \emph{cross-order interaction}. 

STI introduces only a small computational overhead. The additional learnable projection $W_e$ accounts for less than 1\% of the backbone parameter count in the evaluated configurations, and the mechanism requires no dedicated attention heads or additional attention layers. It can therefore be incorporated into a standard transformer sequence while preserving the underlying self-attention operation.


\begin{figure*}[t]
    \centering
    \IfFileExists{Diagrama1Odemir.png}{%
        \includegraphics[width=\textwidth]{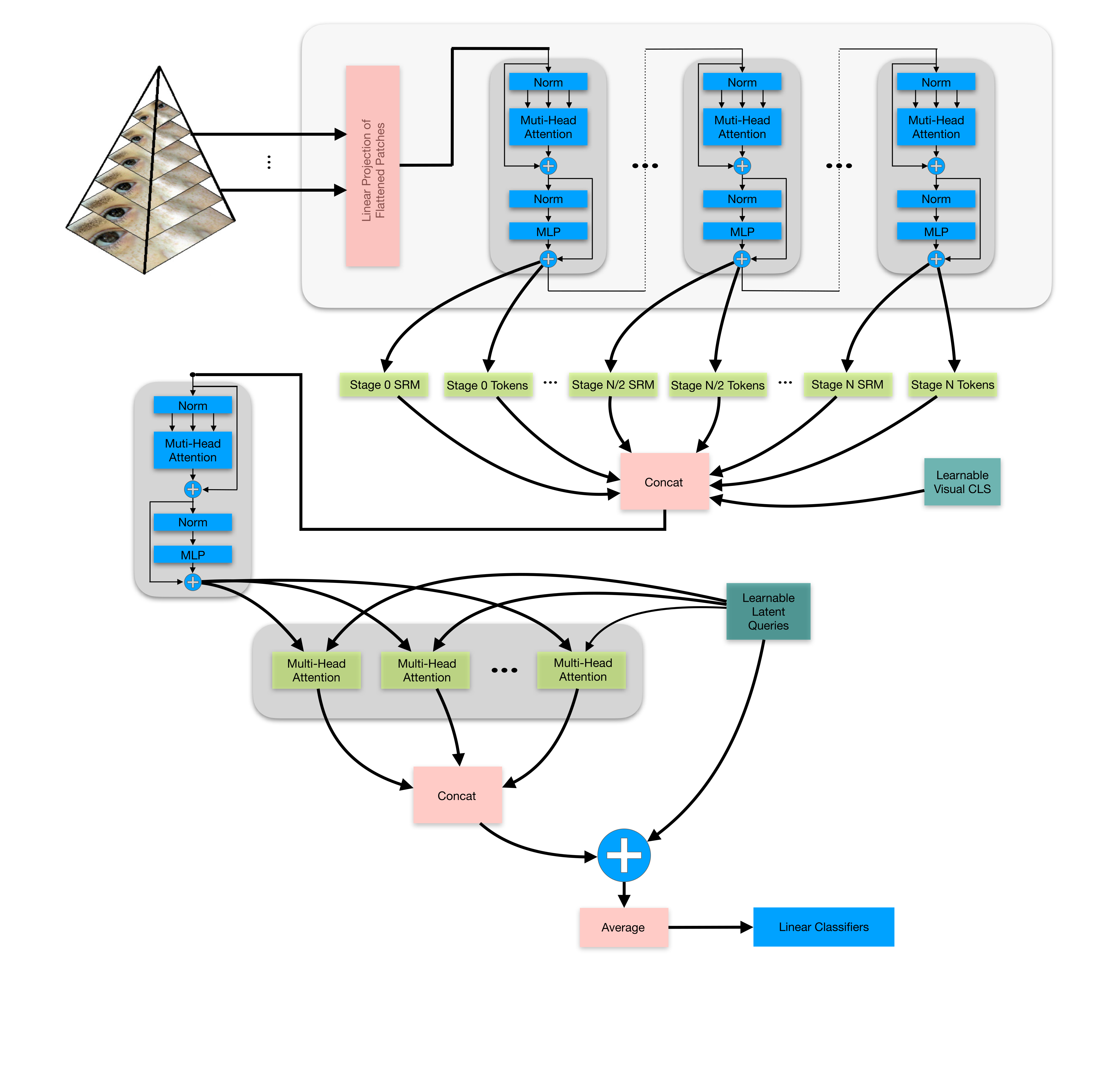}%
    }{%
        \missingfigure{Figures/Diagrama1Odemir.png}%
    }
    \caption{Architecture of the proposed HiPerViT network. The pipeline consists of: (1) a multi-scale ViT backbone for hierarchical feature extraction, (2) SRM projection to embed second-order statistics, (3) fusion through a Latent Transformer producing Key/Value pairs, (4) a Perceiver-style module with learnable latent queries and a self-attention tower for high-order interactions, and (5) final aggregation and classification.}
    \label{fig:full_architecture}
\end{figure*}

\subsection{HiPerViT: Extraction, Interaction, and Distillation}

This subsection details the architecture of \textbf{HiPerViT}, a unified model that integrates hierarchical multi-scale feature extraction from a ViT backbone with second-order feature aggregation via compact bilinear pooling and efficient latent compression.

\subsubsection{Global-Local Input Pathway}

The pipeline begins with a \textbf{Global-Local Dual Pathway}. Texture recognition benefits from analyzing an image at both a broad structural scale and a high-resolution detail scale.

To implement this, and as conceptually depicted by the input section of the architecture in Figure \ref{fig:full_architecture}, the raw input image $\mathbf{I}$ is transformed into two distinct complementary views: a downsampled \emph{global context} image $X_g$, and a high-resolution \emph{local foveal crop} $X_\ell$. This dual-pathway strategy guarantees that the subsequent feature extractor receives data emphasizing different textural scales:
\begin{itemize}
    \item \textbf{Global Context ($X_g$):} Captures overall structural information and long-range dependencies across macro-patterns.
    \item \textbf{Local Foveal View ($X_\ell$):} Preserves fine-grained micro-patterns and high-frequency local statistics.
\end{itemize}

Each view ($X_g$ and $X_\ell$) is then independently forwarded through a \emph{single, shared-weight} ViT backbone; that is, both scales are processed by the same set of parameters rather than independent parallel branches.
This weight-sharing design has two important consequences: (i) memory cost remains proportional to a single backbone, since no additional parameters are introduced for the extra view, and (ii) the shared filters are exposed to the same visual content at multiple scales during training, which encourages the emergence of scale-equivariant internal representations and improves generalization.
The resulting hierarchical feature maps from both views are subsequently aggregated and passed to the Distillation module.

\subsubsection{Hierarchical Feature Extraction and Second-Order Aggregation}

Following the global-local input preparation, the images $X_g$ and $X_\ell$ are processed by the Vision Transformer backbone. We leverage its inherent design to produce a sequence of feature maps at decreasing spatial resolutions, generating a set of Stage Activation Maps (SAMs) from distinct stages of the hierarchy.

The feature extraction step selects a subset of these SAMs, $\mathbf{F}_{s_i}$, corresponding to different depths $s_i$. These selected feature maps are first tokenized and transformed into sequences $\mathbf{T}_{s_i}$.

From this point, the architecture diverges into two parallel streams of representation:

\begin{enumerate}
    \item \textbf{Token Stream ($\mathbf{T}_{multi}$):} The individual stage tokens $\mathbf{T}_{s_i}$ are concatenated along the sequence dimension to form a single, aggregated multi-scale token object, $\mathbf{T}_{multi}$:
    \begin{equation}
    \mathbf{T}_{multi} = \text{Concat}(\mathbf{T}_{s_1}, \mathbf{T}_{s_2}, \dots, \mathbf{T}_{s_k})
    \end{equation}
    where $\mathbf{T}_{multi} \in \mathbb{R}^{N \times D}$, and $N = \sum_{i=1}^{k} H_i W_i$. This unified set serves as the high-dimensional input for the subsequent attention-based Distillation module.
    
    \item \textbf{Second-Order Stream ($\mathbf{Z}_{SRM}$):} The Second-order Representation Module (SRM) is applied independently to the tokens of each stage $\mathbf{T}_{s_i}$. Conceptually, this module aims to capture the full second-order statistics of the features, which corresponds to the sum of outer products of the token vectors:
    \begin{equation}
    \mathbf{Z}_{s_i}^{\text{full}} = \sum_{j=1}^{H_i W_i} \mathbf{t}_{j} \mathbf{t}_{j}^\top
    \end{equation}
    However, directly computing this high-dimensional tensor is computationally prohibitive. Therefore, in our implementation (detailed in Sec.~\ref{sec:srm_implementation}), we approximate this operation using Compact Bilinear Pooling (CBP) via the Tensor Sketch trick, generating a compact stage-specific descriptor $\mathbf{Z}_{s_i}$.
    These stage-specific second-order representations, $\mathbf{Z}_{s_i}$, are then concatenated to form the final, compact multi-scale second-order feature vector $\mathbf{Z}_{SRM}$:
    \begin{equation}
    \mathbf{Z}_{SRM} = \text{Concat}(\mathbf{Z}_{s_1}, \mathbf{Z}_{s_2}, \dots, \mathbf{Z}_{s_k})
    \end{equation}
\end{enumerate}

The aggregated token set $\mathbf{T}_{multi}$ and the combined second-order vector $\mathbf{Z}_{SRM}$ are both passed forward to the next stage, ensuring the system integrates both sequential and correlational feature streams.

\subsubsection{Early Interaction, Latent Distillation, and Classification}

The final phase of the HiPerViT pipeline integrates first-order spatial tokens with second-order statistical summaries via cross-order self-attention, followed by an asymmetric distillation. We describe the default \emph{interact$\rightarrow$distill} ordering here; Sec.~\ref{sec:topology_robustness} shows that alternative orderings yield statistically equivalent performance.

\textbf{Cross-Order Visual Encoding:}
The high-dimensional spatial token sequence $\mathbf{T}_{multi}$ and the compact multi-scale second-order vector $\mathbf{Z}_{SRM}$ are not simply concatenated at the output. Instead, $\mathbf{Z}_{SRM}$ is embedded into the same dimensional space as the spatial tokens to form a \emph{statistical token}, denoted $\mathbf{T}_{srm}$. An augmented visual sequence is formed by prepending a class token and the statistical token to the multiscale spatial tokens:
\begin{equation}
\mathbf{Z}_{vis} = \text{Concat}(\text{[CLS]}, \mathbf{T}_{srm}, \mathbf{T}_{multi})
\end{equation}
This sequence is processed by a lightweight visual Transformer encoder, facilitating \emph{cross-order and cross-scale self-attention}. The spatial tokens dynamically attend to the global statistical token (and vice-versa), integrating fine-grained correlations before any dimensionality reduction occurs.

\textbf{Feature Distillation (The Perceiver-style Latent Bottleneck):}
The enriched sequence $\widetilde{\mathbf{Z}}_{vis} \in \mathbb{R}^{(N+2) \times D}$ must still be condensed. This is achieved through a Perceiver-style latent bottleneck. A set of trainable, low-dimensional Latent Query tokens $\mathbf{L} \in \mathbb{R}^{B \times D'}$ (where $B \ll N$) uses asymmetric cross-attention to query the interacting visual stream:
\begin{equation}
\mathbf{L}_{distilled} = \text{CrossAttention}(\mathbf{L} \text{ as Query}, \widetilde{\mathbf{Z}}_{vis} \text{ as Key/Value})
\end{equation}
In the default configuration, HiPerViT distills a sequence where first- and second-order features have already contextually modulated each other. However, controlled topology ablations (Sec.~\ref{sec:topology_robustness}) demonstrate that reversing this ordering or deferring SRM integration to a late-fusion stage yields equivalent accuracy, confirming that the gain arises from the SRM statistical prior rather than a specific wiring.

\textbf{Classification:}
The distilled latent representation $\mathbf{L}_{distilled}$ is aggregated via mean pooling to produce a final, compact vector:
\begin{equation}
\mathbf{V}_{final} = \frac{1}{L} \sum_{i=1}^L \mathbf{L}_{distilled}^{(i)}
\end{equation}
This ultimate feature vector $\mathbf{V}_{final}$ is fed into a linear classification head to project onto the number of target classes $C$:
\begin{equation}
\hat{y} = \text{Softmax}(\text{MLP}(\mathbf{V}_{final}))
\end{equation}
The network is trained end-to-end by minimizing the standard Cross-Entropy loss $\mathcal{L}_{\text{CE}}$ between the predicted probabilities $\hat{y}$ and the ground-truth labels $y$.

\subsection{Overview and Notation}

Given a batch of images
\begin{equation}
    \mathcal{X} = \{ X^{(b)} \}_{b=1}^B, \qquad X^{(b)} \in \mathbb{R}^{3 \times H \times W},
\end{equation}
we generate two complementary views: a \emph{global} image $X_g$ and a set of \emph{local} crops $X_\ell$, mimicking peripheral and foveal vision. Each image is tokenized by a Vision Transformer (ViT) backbone into patch embeddings:
\begin{equation}
    \mathbf{Z}_g \in \mathbb{R}^{B \times n_g \times D}, \qquad
    \mathbf{Z}_\ell \in \mathbb{R}^{B \times n_\ell \times D},
\end{equation}
where $D$ is the embedding dimension and $n_g$, $n_\ell$ are the number of tokens for global and local views, respectively.

\subsection{Second-Order Statistical Branch}\label{sec:srm_implementation}

Textures are defined not only by first-order statistics (mean activations) but also by \emph{correlations} among features. Inspired by this, we add a compact bilinear pooling module that approximates the outer product $\mathbf{f}\mathbf{f}^\top$ of a feature vector $\mathbf{f}\in\mathbb{R}^C$ without quadratic cost. 

The Count Sketch projection $S:\mathbb{R}^C\to \mathbb{R}^M$ is defined as
\begin{equation}
    S(\mathbf{f})[m] = \sum_{i=1}^C s_i f_i \, \mathbf{1}_{h_i=m},
\end{equation}
where $h \in \{1,\dots,M\}^C$ and $s \in \{\pm 1\}^C$ are fixed random vectors. The bilinear feature is then sketched using convolution in the Fourier domain:
\begin{equation}
    \widehat{S(\mathbf{f} \otimes \mathbf{f})} = \widehat{S(\mathbf{f})} \odot \widehat{S(\mathbf{f})},
\end{equation}
with $\odot$ denoting elementwise multiplication and $\widehat{\cdot}$ the FFT. The inverse FFT yields the compact bilinear feature $\mathbf{c}\in\mathbb{R}^M$, normalized via signed square-root and $\ell_2$:
\begin{equation}
    \mathbf{z}_{\text{srm}} = 
    \frac{\mathrm{sign}(\mathbf{c}) \cdot \sqrt{|\mathbf{c}| + \varepsilon}}
         {\| \mathrm{sign}(\mathbf{c}) \cdot \sqrt{|\mathbf{c}| + \varepsilon}\|_2}.
\end{equation}

This vector $\mathbf{z}_{\text{srm}}$ forms a \emph{statistical token}, directly encoding second-order co-occurrences. Neuroscience suggests such pairwise coding is crucial in mid-level vision \cite{Freeman2013}, providing a biological rationale for this branch.

\subsection{Projection and Cross-Order Visual Encoding}

A defining structural characteristic of HiPerViT is the shared interaction space for heterogeneous features. We project all branches into a unified embedding space of dimension $D$:
\begin{align}
    \mathbf{p}_g &= \mathbf{Z}_g W_p + b_p, \\
    \mathbf{p}_\ell &= \mathbf{Z}_\ell W_p + b_p, \\
    \mathbf{p}_{\text{srm}} &= \mathbf{z}_{\text{srm}} W_s + b_s,
\end{align}
with $W_p\in\mathbb{R}^{D \times D}$ and $W_s\in\mathbb{R}^{M \times D}$. By embedding $\mathbf{z}_{\text{srm}}$ as $\mathbf{p}_{\text{srm}}$, we cast the dense second-order summary as a \emph{statistical token}. We then concatenate these tokens into a single sequence:
\begin{equation}
    \mathbf{Z}_{\text{vis}} = [\text{[CLS]}, \mathbf{p}_{\text{srm}}, \mathbf{p}_g, \mathbf{p}_\ell] \in \mathbb{R}^{B \times (1 + 1 + n_g + n_\ell) \times D},
\end{equation}
and refine them with a lightweight Transformer encoder:
\begin{equation}
    \widetilde{\mathbf{Z}}_{\text{vis}} = \mathrm{TransEnc}(\mathbf{Z}_{\text{vis}}).
\end{equation}
This explicitly models mutual dependencies: the spatial tokens ($\mathbf{p}_g, \mathbf{p}_\ell$) query the global co-occurrence statistics ($\mathbf{p}_{\text{srm}}$) to weight relevant textural patterns, while the statistical token is contextually modulated by the dominant localized structures in the image.

\subsection{Latent Distillation via Cross-Attention}

The representation $\widetilde{\mathbf{Z}}_{\text{vis}}$ remains long and redundant. To distill information, we introduce $L$ learnable latent vectors $\mathbf{Q}\in \mathbb{R}^{L \times D}$, updated by cross-attention:
\begin{equation}
    \mathbf{L} = \mathrm{softmax}\!\left(\frac{\mathbf{Q}\widetilde{\mathbf{Z}}_{\text{vis}}^\top}{\sqrt{D}}\right)\widetilde{\mathbf{Z}}_{\text{vis}}.
\end{equation}
The output $\mathbf{L}\in\mathbb{R}^{B\times L\times D}$ compresses hundreds of tokens into $L \ll N$ informative latents, condensing distributed multi-scale signals into compact codes for recognition.

The number of latent queries $L$ is a critical hyperparameter governing the trade-off between compression and representational fidelity. A small $L$ forces the model to abstract away high-frequency spatial details, favoring dominant macro-structures and global context. Conversely, a larger $L$ preserves more fine-grained micro-textural information but increases the computational cost of the subsequent self-attention layers. Empirically, we find that a moderate $L$ (e.g., 64--128) suffices to capture the relevant texture statistics without retaining pixel-perfect spatial reconstruction, effectively filtering out high-frequency noise while preserving the structural signatures required for classification.

\subsection{Classification}

We aggregate latents by mean pooling:
\begin{equation}
    \mathbf{h} = \frac{1}{L} \sum_{i=1}^L \mathbf{L}_i,
\end{equation}
and predict with a linear classifier:
\begin{equation}
    \hat{\mathbf{y}} = \mathrm{softmax}(W_c^\top \mathbf{h} + b_c),
\end{equation}
with $W_c\in\mathbb{R}^{D\times C}$ and $C$ the number of classes.

\subsection{Training Methodology}

We train with \emph{Soft Target Cross-Entropy}, compatible with mixup:
\begin{equation}
    \mathcal{L}(\hat{\mathbf{y}}, \tilde{\mathbf{t}}) 
    = -\sum_{c=1}^C \tilde{t}_c \log \hat{y}_c,
\end{equation}
where $\tilde{\mathbf{t}} = \lambda \mathbf{t}_a + (1-\lambda)\mathbf{t}_b$ is a convex combination of one-hot labels. We further apply RandAugment and Layer-wise Learning Rate Decay (LLRD):
\begin{equation}
    \eta_i = \eta_0 \cdot \rho^{S-1-i},
\end{equation}
where $i$ indexes blocks of the backbone, $S$ is the total depth, and $\rho<1$ the decay factor. This schedule applies lower learning rates to early, well-pretrained layers while allowing higher layers to adapt quickly.

\subsection{Architectural Synthesis}

A key consequence of HiPerViT's design emerges from the role of the statistical token within the fusion pipeline.
By making second-order statistics available as explicit tokens to cross-order self-attention, the model decouples final discriminability from the intrinsic linear separability of the SRM vector itself.
Cross-order self-attention enables the spatial token stream to conditionally re-weight second-order correlations based on contextual structure: when the SRM signal is highly informative, the attention mechanism amplifies it; when it is noisy or under-determined, the spatial tokens carry the discriminative burden.
Consequently, the system's classification power becomes a property of \emph{token interaction} rather than of standalone feature quality.
This reframes compact bilinear pooling, traditionally viewed as a static, orderless descriptor appended at the output of a network, into a dynamically contextualized representation mechanism whose utility is determined by the \emph{presence} of cross-order attention rather than by a specific interaction--distillation ordering or the extraction depth of the embedding.
As we demonstrate empirically, this property manifests as two complementary invariances: (i)~final accuracy is invariant to the backbone depth at which the SRM is computed, even when raw discriminability varies by more than 30 percentage points (Sec.~\ref{sec:mechanistic}), and (ii)~final accuracy is invariant to whether the statistical token interacts before or after latent distillation (Sec.~\ref{sec:topology_robustness}).

\section{Experiments and Analysis}
\label{sec:experiments}
We conducted a comprehensive set of experiments to validate the Hierarchical Perceiver--Bilinear Vision Transformers (HiPerViT). Our study is organized around five central questions: (1) How does HiPerViT compare with state-of-the-art (SOTA) texture recognition methods across diverse benchmarks? (2) What is the empirical justification for our architectural choices, particularly multi-scale input and hierarchical feature fusion? (3) How do second-order statistics evolve across backbone depth, and what does this reveal about HiPerViT's robustness to the representational source of its SRM branch? (4) Does the interact$\rightarrow$distill fusion preserve generic representation transferability or induce task-specific specialization? (5) How does HiPerViT balance accuracy with computational and memory cost across different ViT backbones?

\subsection{Experimental Setup}
\textbf{Datasets and Evaluation Protocol.} We evaluate HiPerViT on six public texture benchmarks with varying sizes, complexity, and acquisition conditions: DTD \cite{DTD2014}, FMD \cite{FlickrMaterialDatabase}, KTH-TIPS2-b \cite{Mallikarjuna2006}, GTOS-Mobile, USPTex \cite{USPTex}, and 1200Tex \cite{Casanova2009}. This selection captures both controlled and in-the-wild conditions, enabling a robust assessment of generalization. To ensure rigorous and replicable evaluation, we follow established splitting protocols: for DTD, results are averaged over the 10 official train/test partitions; for 1200Tex, we apply a Repeated Stratified 5-Fold Cross-Validation (2 repeats); for all other datasets, we adopt the standard 5-fold cross-validation or predefined train/test splits as provided by the original authors.

\textbf{Implementation Details and Optimization.} The multi-scale input pathway generates a \emph{global} view resized to $448 \times 448$ pixels via bicubic interpolation and a \emph{local} foveal view derived from a Random Resized Crop (scale $0.5$ to $1.0$) of size $224 \times 224$ during training, transitioning to a deterministic center crop for evaluation. The ViT backbones (Small, Base, and Large) are exclusively initialized using DINOv2 self-supervised weights \cite{Oquab2024}, ensuring no text-supervision is leaked into our vision-only evaluation. 
We train the model using the AdamW optimizer with a base learning rate of $3 \times 10^{-5}$ and a weight decay of $0.05$. Optimization proceeds for 10 epochs with an effective batch size of 32 (via gradient accumulation), following a cosine annealing schedule with a 10\% linear warmup phase. To preserve the pre-trained structural representations, we apply Layer-wise Learning Rate Decay (LLRD) with a decay factor of $0.8$ per block from the top down, while the randomly initialized classifier head and Perceiver bottleneck receive a $10\times$ learning rate multiplier ($3 \times 10^{-4}$). 

\textbf{Augmentation Strategy.} The network is trained with a Soft Target Cross-Entropy loss compatible with mixed augmentations. For unbalanced datasets, explicit class weighting is applied. Specifically, we apply RandAugment with magnitude $(4, 12)$ strictly to the global view. In addition, we apply a stochastic mixing strategy per batch where CutMix ($\alpha=1.0$) is selected with 50\% probability; otherwise, standard Mixup ($\alpha=0.5$) is applied. During evaluation, we report results using Test-Time Augmentation (TTA), averaging the softmax predictions over the original image and horizontal, vertical, and $180^\circ$ flipped spatial variations.

\subsection{Comparison with State-of-the-Art}

Table~\ref{tab:combined_sota} consolidates our comparison against the strongest previously published vision-only methods and transformer-based architectures. The unified structure, divided into subtables (a) and (b), enables a clearer side-by-side interpretation of both external SOTA results and controlled transformer baselines.

It is important to contextualize the metrics in these two subtables. The "Previous SOTA" results in Table~\ref{tab:combined_sota}a reflect values exactly as reported in the respective original publications. As such, these numbers are inherently heterogeneous: they stem from distinct underlying architectures (e.g., ConvNeXt-XXL vs. ViT-L), varying pre-training modalities, and highly specific hyperparameter tuning efforts tailored by the original authors. 

To provide a more rigorous, apples-to-apples evaluation of our architectural claims, Table~\ref{tab:combined_sota}b compares HiPerViT against standard transformer architectures trained under the \emph{identical protocol} detailed in Section 5.1. While this protocol establishes a strong, modern baseline anchored around ViT and DINOv2 weights, we acknowledge that a single, unified training recipe may not represent the absolute optimal hyperparameter configuration for every competing architecture (such as Swin or ConvNeXt). Nevertheless, holding the training regime, data splits, and augmentations perfectly constant isolates the performance delta attributable strictly to HiPerViT's hierarchical fusion and second-order pooling mechanisms.

Overall, HiPerViT establishes new vision-only state-of-the-art on all six benchmarks, with substantial gains on five. On DTD, our model surpasses a strong ViT-L/14 baseline by +3.05\,pp, illustrating the benefit of hierarchical fusion on highly variable textures. The largest improvements relative to previously published results appear on GTOS-Mobile (+10.48\,pp over RADAM~\cite{Scabini2023}) and 1200Tex (+10.1\,pp over VisGraphNet~\cite{Florindo2021}), demonstrating robustness under challenging illumination and high-resolution settings. On USPTex, HiPerViT achieves perfect accuracy (100.00\%), surpassing the previous best (99.82\%). However, rather than simply claiming an architectural breakthrough, this perfect score on USPTex signals severe dataset saturation. It suggests that the discriminative power of modern DINOv2 pre-trained features—potentially combined with structural artifacts of the dataset itself, such as highly similar or near-duplicate textural crop samples across splits or preprocessing leakage—is sufficient to entirely over-solve the benchmark.

When compared to alternative transformer architectures trained with identical protocols (Table~\ref{tab:combined_sota}b), HiPerViT outperforms all baselines across all datasets, confirming the effectiveness of its multiscale fusion and hierarchical feature integration.

\begin{table*}[t]
\centering
\footnotesize
\renewcommand{\arraystretch}{1.1}
\caption{Top-1 accuracy (\%) across six texture benchmarks.
(a) Comparison with previously published vision-only state-of-the-art.
(b) Comparison with transformer-based models trained under identical settings.
Best results are in \textbf{bold}.}
\label{tab:combined_sota}

\textbf{(a) Comparison with previously published vision-only SOTA.}
\label{tab:sota_comparison}

\vspace{0.3em}
\begin{tabular}{@{}lccr@{}}
\toprule
\textbf{Dataset} & \textbf{HiPerViT (\%)} & \textbf{Previous SOTA} & \textbf{Accuracy (\%)} \\
\midrule
DTD        & \textbf{93.05} & Linear-FT (ViT-L/14)       & 90.00 \\
FMD        & \textbf{97.80} & RADAM (ConvNeXt-XXL)       & 97.10 \\
KTH-TIPS2-b& \textbf{98.42} & RADAM (ConvNeXt-XXL)       & 97.40 \\
GTOS-Mobile& \textbf{94.58} & RADAM (ConvNeXt-B)         & 84.10 \\
USPTex     & \textbf{100.00}& Local Graphs via Random Encoding & 99.82 \\
1200Tex    & \textbf{97.50} & VisGraphNet                & 87.40 \\
\bottomrule
\end{tabular}

\vspace{0.5em}

\textbf{(b) Comparison with transformer architectures under identical training settings.}

\vspace{0.3em}
\begin{tabular}{@{}lcccccc@{}}
\toprule
\textbf{Model} & \textbf{DTD} & \textbf{FMD} & \textbf{KTH-TIPS2-b} & \textbf{GTOS-Mobile} & \textbf{USPTex} & \textbf{1200Tex} \\
\midrule
HiPerViT & \textbf{93.05} & \textbf{97.80} & \textbf{98.42} & \textbf{94.58} & \textbf{100.00} & \textbf{97.50} \\
ViT-B/16 & 73.40 & 91.50 & 85.86 & 84.72 & 99.78 & 97.00 \\
DeiT     & 69.73 & 89.00 & 82.32 & 88.79 & 99.78 & \textbf{97.50} \\
Swin     & 77.34 & 92.00 & 88.89 & 89.20 & \textbf{100.00} & 97.00 \\
VORTEX   & 87.10 & 93.90 & 95.10 & 93.80 & \textbf{100.00} & \textbf{97.50} \\
\bottomrule
\end{tabular}

\vspace{0.4em}
\begin{minipage}{0.98\textwidth}
\footnotesize
\textit{Sources for panel (a):} Linear-FT (ViT-L/14) \cite{Ortiz2023}; RADAM \cite{Scabini2023}; Local Graphs via Random Encoding \cite{Fares2025}; VisGraphNet \cite{Florindo2021}. Dataset references: DTD \cite{DTD2014}, FMD \cite{FlickrMaterialDatabase}, KTH-TIPS2-b \cite{Mallikarjuna2006}, USPTex \cite{USPTex}, 1200Tex \cite{Casanova2009}.
\end{minipage}
\end{table*}

\subsection{Impact of Multi-Scale and Hierarchical Fusion}

\begin{figure}[h!]
    \centering
    \IfFileExists{1_performance_plane_scales.png}{%
        \includegraphics[width=\linewidth]{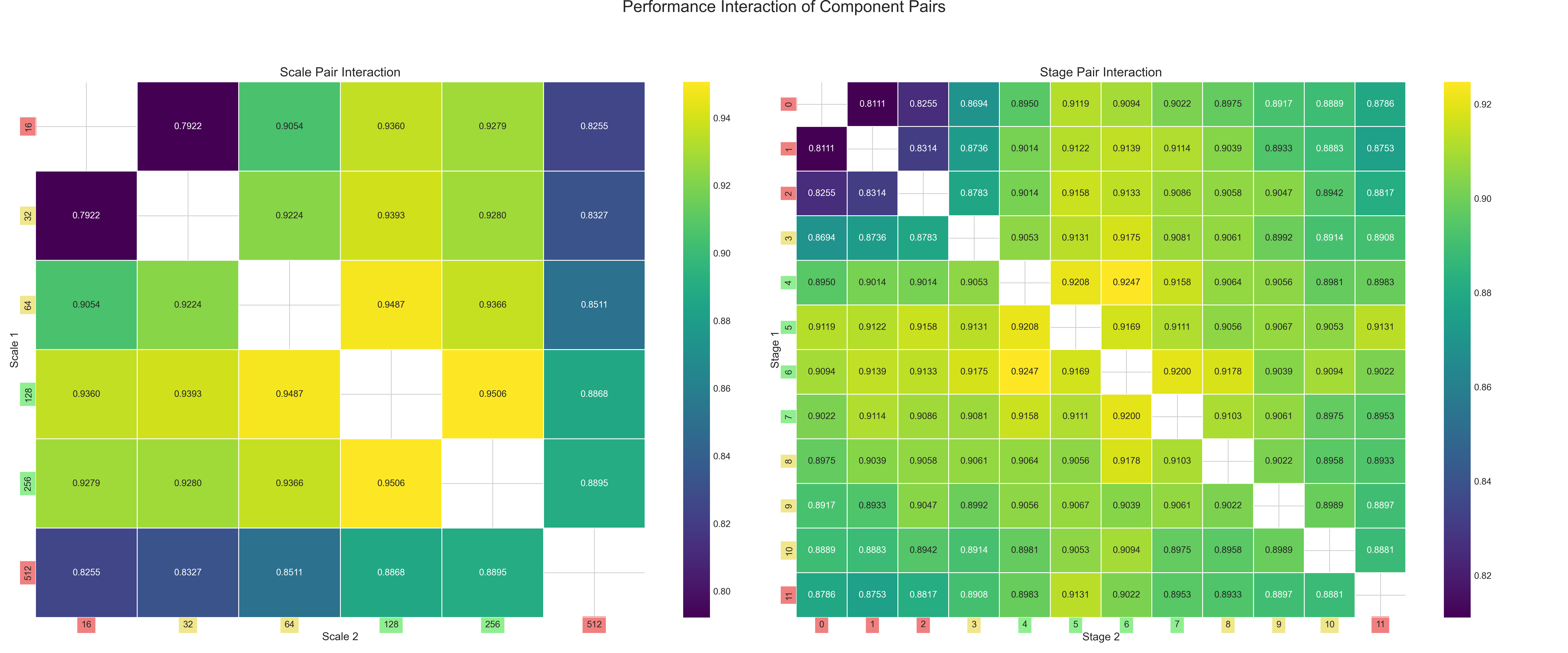}%
    }{%
        \missingfigure{Figures/1_performance_plane_scales.png}%
    }
    \caption{Performance heatmaps for input scale pairs (left) and backbone stage pairs (right). Cells show mean accuracy. Axis labels are color-coded by individual component performance tier.}
    \label{fig:interaction_heatmaps}
\end{figure}

To disentangle design choices, we performed a hyperparameter sweep over input scale pairs and backbone stage pairs. Results are summarized in Figure \ref{fig:interaction_heatmaps}.

Two clear trends emerge: (i) Combining coarse and fine input scales (e.g., 256px and 128px) consistently outperforms single-scale setups, validating the need for complementary global and local context. (ii) Fusing early- or mid-level backbone features with deeper semantic stages yields the best results, outperforming shallow-only or deep-only combinations. This confirms our core claim: HiPerViT succeeds not by depth alone, but by coupling diverse representational levels.

The observation that intermediate backbone stages paired with deeper layers maximize accuracy is consistent with the principle that mid-level integration of local detail and coarse context precedes effective high-level recognition.

\begin{table}[t]
    \centering
    \caption{Top-performing configurations of scale and stage combinations.}
    \label{tab:top_combinations}
    \begin{tabular}{@{}cccc@{}}
    \toprule
    \textbf{Rank} & \textbf{Scales} & \textbf{Stages} & \textbf{Accuracy (\%)} \\
    \midrule
    1 & 128-64 & 1-5 & 97.50 \\
    2 & 256-128 & 1-5 & 97.50 \\
    3 & 256-64 & 8-9 & 97.08 \\
    4 & 256-128 & 1-8 & 97.08 \\
    5 & 256-128 & 4-5 & 97.08 \\
    ... & ... & ... & ... \\
    \bottomrule
    \end{tabular}
\end{table}

\subsection{Component Ablation: Contribution of Each Module}
\label{sec:ablation_components}

To isolate the contribution of each architectural component in HiPerViT, we perform a controlled ablation study over the same training protocol, data preprocessing, and evaluation procedure used throughout Sec.~\ref{sec:experiments}. We progressively disable or replace individual modules while keeping the remaining pipeline unchanged. In particular, we consider: (i) a plain transformer backbone (\emph{Backbone Only}); (ii) feeding only the multi-scale image pyramid without hierarchical fusion (\emph{Multi-Scale Image Only}); (iii) using only the SRM branch (\emph{SRM Only}); (iv) using only the Perceiver-based bottleneck pathway (\emph{Perceiver Only}); and (v) the full HiPerViT model. Results (Top-1 accuracy, \%) are summarized in Tab.~\ref{tab:ablation_components}.

\vspace{0.25em}
\noindent\textbf{Impact on controlled texture benchmarks.}
Across DTD, 1200Tex, USPTex, and FMD, the full model consistently outperforms the strongest partial variants, indicating that the gains do not stem from a single isolated component but from their combination. Notably, the \emph{Perceiver Only} variant produces a substantial improvement over the backbone on DTD (from 83.35 to 91.44), suggesting that the bottleneck aggregation mechanism is particularly effective at consolidating cues relevant to diverse texture patterns. However, on 1200Tex and USPTex the backbone alone already achieves near-perfect performance. The \emph{Multi-Scale Image Only} variant reaches exactly 100.00\% on USPTex. The fact that a partial pipeline can perfectly solve USPTex corroborates the notion that this dataset provides limited granular signal for evaluating complex architectural additions, as the task is trivialized by the baseline features and multi-scale pre-processing. The improvements from the full HiPerViT assembly are therefore most reliably observed on challenging benchmarks with robust performance headroom, like DTD and 1200Tex.

\vspace{0.25em}
\noindent\textbf{Role of SRM and multi-scale cues.}
The SRM-only configuration improves over the backbone on KTH-TIPS2-b (from 87.79 to 89.23), which is consistent with the hypothesis that high-frequency residual cues can be informative under certain material/illumination changes. By contrast, relying solely on multi-scale imagery without hierarchical fusion does not yield consistent gains and can slightly degrade performance on KTH-TIPS2-b, indicating that scale diversity alone is insufficient unless it is integrated via a structured fusion mechanism.

\vspace{0.25em}
\noindent\textbf{Complementarity and full-model behavior.}
The full HiPerViT configuration achieves the best overall results on three of the four fully reported datasets (DTD, 1200Tex, FMD), while remaining competitive on USPTex and KTH-TIPS2-b. This pattern supports the central design premise of HiPerViT: \emph{multi-scale inputs}, \emph{SRM-guided residual cues}, and \emph{Perceiver-style global aggregation} are complementary, and the hierarchical fusion strategy is essential for translating this complementarity into consistent accuracy gains across benchmarks. In full-data settings the effect of the statistical token is modest, but becomes substantially larger under low-data training (Sec.~\ref{sec:lowdata}).

\vspace{0.25em}
\noindent\textbf{Capacity-Dependence of Ablation Trends (ViT-S vs. ViT-L).}
It should be noted that while our headline state-of-the-art results (Table \ref{tab:combined_sota}) are achieved using the highest-capacity \textbf{ViT-L/16} backbone, this ablation study is purposefully conducted using the smaller \textbf{ViT-S/16} backbone. This deliberate discrepancy serves a structural aim: immense parameter models often possess the raw capacity to implicitly approximate complex visual patterns, which can mask the specific contributions of explicit structural priors like multi-scale fusion or SRM texture encoding. By restricting the study to a lower-capacity backbone, we isolate the architectural impact of the HiPerViT modules, demonstrating that the structural topology itself—and not just parameter count—drives the recognition improvements. While the functional trends of complementarity hold across all scales, the relative magnitude of improvement provided by our modules is naturally more pronounced in compact backbones that rely heavily on these explicit priors.

\begin{table}[t]
\centering
\caption{Component ablation of HiPerViT using the ViT-S/16 backbone (Top-1 accuracy, \%). The full model differs from Table~\ref{tab:combined_sota} because this ablation uses ViT-S/16, whereas the main comparison reports the best ViT-L/16 configuration. Best results per column are in bold.}
\label{tab:ablation_components}

\scriptsize
\setlength{\tabcolsep}{2.2pt}
\renewcommand{\arraystretch}{1.05}

\begin{tabular}{@{}lcccccc@{}}
\toprule
\textbf{Variant} &
\textbf{DTD} &
\textbf{1200} &
\textbf{USP} &
\textbf{FMD} &
\textbf{KTH} &
\textbf{GTOS} \\
\midrule
Backbone
& 83.35 & 96.97 & 99.74 & 95.05 & 87.79 & 93.42 \\

Multi-Scale
& 82.98 & 94.17 & \textbf{100.00} & 95.05 & 83.42 & 91.48 \\

SRM
& 82.98 & 95.00 & 99.74 & 92.08 & \textbf{89.23} & 93.65 \\

Perceiver
& 91.44 & 94.17 & 98.95 & 96.04 & 88.64 & \textbf{94.92} \\

HiPerViT
& \textbf{92.66} & \textbf{97.50} & 99.74 &
\textbf{97.03} & 88.39 & 93.04 \\
\bottomrule
\end{tabular}

\end{table}

\subsubsection{Data-Efficiency of the Statistical Token}
\label{sec:lowdata}

Does the SRM statistical token provide larger benefits when supervision is scarce, i.e., when the backbone cannot easily internalize co-occurrence priors from data?
To test this, we subsample the DTD training set using stratified sampling at \{10, 20, 50, 100\}\% of the training data, keeping the test set unchanged.
We train three variants under the identical recipe described in Sec.~5.1:
(i)~\textbf{SRM token} (full HiPerViT),
(ii)~\textbf{GAP token} (first-order pooled descriptor injected as the statistical token), and
(iii)~\textbf{none} (no statistical token; Perceiver-only fusion).
We report mean$\pm$std over 3 seeds with TTA.

\begin{table}[t]
\centering
\caption{Low-data regime on DTD. SRM, GAP, and None variants were trained under identical conditions at each fraction. $\Delta$ columns report the advantage of SRM over the comparison variant. Values are mean$\pm$std over 3 seeds.}
\label{tab:lowdata}

\footnotesize
\setlength{\tabcolsep}{3pt}
\renewcommand{\arraystretch}{1.06}

\begin{tabular}{@{}lccccc@{}}
\toprule
\textbf{Frac.} &
\textbf{SRM} &
\textbf{GAP} &
\textbf{None} &
\textbf{$\Delta$(S$-$N)} &
\textbf{$\Delta$(S$-$G)} \\
\midrule
10\%  & 81.52$\pm$0.44 & 80.90$\pm$0.43 & 79.88$\pm$0.26 & +1.64 & +0.62 \\
20\%  & 88.55$\pm$0.25 & 87.71$\pm$0.90 & 87.94$\pm$0.33 & +0.61 & +0.84 \\
50\%  & 91.45$\pm$0.29 & 91.58$\pm$0.09 & 91.29$\pm$0.58 & +0.16 & $-0.13$ \\
100\% & 92.77$\pm$0.13 & 92.57$\pm$0.09 & 92.34$\pm$0.37 & +0.43 & +0.20 \\
\bottomrule
\end{tabular}

\end{table}

The SRM token yields its largest gains in the low-data regime: at 10\% of DTD training data, SRM improves over the no-token variant by $+$1.64\,pp, while the advantage narrows as the fraction increases and becomes marginal at 50--100\%.
This trend suggests that second-order co-occurrence modeling primarily acts as an inductive bias that compensates for limited supervision;
with sufficient data and a strong pretrained backbone, the fusion tower can recover comparable performance using first-order summaries or no explicit statistical token.
Notably, at 50\% data the GAP token slightly outperforms SRM ($-$0.13\,pp), underscoring that the benefit is regime-dependent rather than absolute.

This complements the mechanistic analysis in Sec.~\ref{sec:mechanistic}: while the interact$\rightarrow$distill topology is robust to the origin and quality of second-order cues, explicit SRM modeling provides measurable benefits precisely when learning signals are scarce.

\subsection{Fusion Topology Robustness and Order Invariance}
\label{sec:topology_robustness}

A natural question is whether the performance of HiPerViT depends on the specific ordering of interaction and distillation, or whether the gain stems primarily from the SRM statistical prior itself.
To test this, we evaluate three controlled wiring variants that differ only in how the statistical token is integrated with the spatial token stream, while matching parameter counts and FLOPs:
\begin{itemize}
    \item \textbf{Topology A} ($\mathcal{I}\!\rightarrow\!\mathcal{D}$): Interaction before distillation. The statistical token participates in cross-order self-attention with the full spatial token sequence \emph{before} Perceiver-style compression (the default HiPerViT configuration).
    \item \textbf{Topology B} ($\mathcal{D}\!\rightarrow\!\mathcal{I}$): Distillation before interaction. Spatial tokens are first compressed via the Perceiver bottleneck; the statistical token then interacts with the \emph{distilled} latent sequence.
    \item \textbf{Topology C} (Late fusion): No explicit cross-order attention. The SRM descriptor is concatenated with the pooled latent representation at the classifier input.
\end{itemize}

\begin{table}[t]
\centering
\caption{Fusion topology comparison on DTD (Partition~1, ViT-S/14 DINOv2). Mean $\pm$ std over 3 seeds. $\Delta$ is computed relative to Topology~A under TTA. All topologies are parameter-matched.}
\label{tab:topology_robustness}
\setlength{\tabcolsep}{4pt}
\renewcommand{\arraystretch}{1.1}
\begin{tabular}{lcccc}
\toprule
\textbf{Topology} & \textbf{Acc (no TTA)} & \textbf{Acc (TTA)} & $\boldsymbol{\Delta}$ \textbf{vs A} \\
\midrule
A ($\mathcal{I}\!\rightarrow\!\mathcal{D}$) & 87.11 $\pm$ 0.27 & 87.22 $\pm$ 0.30 & 0.00 \\
B ($\mathcal{D}\!\rightarrow\!\mathcal{I}$) & 87.29 $\pm$ 0.29 & 87.32 $\pm$ 0.35 & $+$0.11 \\
C (Late fusion) & 87.36 $\pm$ 0.40 & 87.47 $\pm$ 0.33 & $+$0.25 \\
\bottomrule
\end{tabular}
\end{table}

Table~\ref{tab:topology_robustness} reveals that \textbf{performance is largely invariant to fusion ordering}.
On DTD Partition~1 (3 seeds), all three topologies fall within a narrow band ($\leq 0.25$\,pp under TTA), with differences well within the run-to-run standard deviation.
Notably, the \emph{best} mean accuracy is achieved by late fusion (Topology~C), not by the interact$\rightarrow$distill ordering (Topology~A).

This result carries an important implication: \textbf{the SRM statistical prior is the key ingredient; the interaction topology is secondary}.
All three variants share the same backbone, the same compact bilinear pooling module, and the same training recipe; only the integration topology differs.
The fact that performance is essentially indistinguishable across wirings indicates that the SRM provides a transferable representational prior whose utility is not sensitive to whether it is injected before or after latent compression.
This finding further validates the STI principle: the gain comes from \emph{making second-order statistics available} to the representation, not from a specific architectural wiring.

Given this near-identical performance, we adopt $\mathcal{I}\!\rightarrow\!\mathcal{D}$ (Topology~A) as the \textbf{reference topology} in all subsequent experiments for two reasons:
(i)~it offers a direct token-level interaction site that facilitates mechanistic analysis (Sec.~\ref{sec:mechanistic}), and
(ii)~it provides a unified implementation across datasets without requiring post-hoc concatenation.
This choice is motivated by interpretability and consistency, not by a claimed performance advantage.
Future work will probe whether regimes of extreme data scarcity, heavy domain shift, or fine-grained spatial corruption reveal conditions under which cross-order attention provides a measurable advantage over late fusion.

\subsection{Mechanistic Analysis of Second-Order Representations}
\label{sec:mechanistic}

The component ablation (Sec.~\ref{sec:ablation_components}) establishes that the SRM branch contributes to HiPerViT's performance. A natural follow-up question is: \emph{does the quality or origin of the second-order statistics matter?} Specifically, since the ViT backbone is known to progressively transform representations from low-level primitives to high-level semantics~\cite{Dosovitskiy2021}, one might hypothesize that SRM vectors derived from earlier layers, which preserve finer-grained texture primitives, should yield stronger texture discriminability than those from deeper, more semantic layers.

To test this hypothesis rigorously, we conduct three complementary analyses on the DTD benchmark using a ViT-S/14 DINOv2 backbone with the fusion topology fixed at the \emph{pre\_interact} configuration. In each experiment, we vary only the backbone block that supplies tokens to the SRM module, selecting block indices at approximately 25\%, 50\%, and 100\% of the backbone depth.

\subsubsection{SRM Capture-Depth Ablation}

We first train the full HiPerViT pipeline while varying only which backbone block supplies tokens to the SRM. All other hyperparameters, including fusion topology, augmentation, and optimizer settings, are held constant. Each variant is trained with three random seeds.

\begin{table}[t]
\centering
\caption{SRM capture-depth ablation on DTD (ViT-S/14 DINOv2). Fusion topology is fixed at \emph{pre\_interact}. Mean $\pm$ std over 3 seeds. $\Delta$ and $p$-values are computed against the Late baseline via paired $t$-test.}
\label{tab:srm_depth}
\begin{tabular}{lccc}
\toprule
\textbf{SRM source} & \textbf{Accuracy (\%)} & $\boldsymbol{\Delta}$ \textbf{vs Late} & $\boldsymbol{p}$\textbf{-value} \\
\midrule
SRM@Early (~25\%)  & 82.91 $\pm$ 0.18 & $-$0.07 & 0.75 \\
SRM@Mid (~50\%)    & 82.91 $\pm$ 0.26 & $-$0.07 & 0.82 \\
SRM@Late (~100\%)  & 82.98 $\pm$ 0.17 &  0.00   & ,   \\
SRM@Multi (all)    & 83.09 $\pm$ 0.16 & $+$0.11 & 0.37 \\
\bottomrule
\end{tabular}
\end{table}

Table~\ref{tab:srm_depth} reveals a striking result: \textbf{final classification accuracy is statistically invariant to SRM capture depth}. All differences are $\leq$0.11\,pp, well within the run-to-run standard deviation, and no paired $t$-test reaches significance ($p \geq 0.37$). This finding directly contradicts the hypothesis that early-layer SRM vectors should be superior for texture recognition.

\subsubsection{Linear Probe Analysis}
\label{sec:linear_probe}

The depth invariance of the full model raises a critical question: do SRM vectors at different depths actually carry the same discriminative information, or is the fusion head compensating for differences?

To answer this, we freeze the pretrained DINOv2 backbone, extract SRM vectors at each capture depth for the entire train and test sets, and train a logistic regression classifier on the raw SRM vectors alone, without the fusion head, Perceiver, or fine-tuning.

\begin{table}[t]
\centering
\caption{Linear probe accuracy on DTD using only the frozen SRM vectors at each backbone depth. No fusion head or fine-tuning is applied.}
\label{tab:linear_probe}
\begin{tabular}{lcc}
\toprule
\textbf{SRM source} & \textbf{Linear probe acc (\%)} & $\boldsymbol{\Delta}$ \textbf{vs Late} \\
\midrule
SRM@Early (~25\%)  & 46.17 & $-$33.56 \\
SRM@Mid (~50\%)    & 71.06 & $-$8.67  \\
SRM@Late (~100\%)  & 79.73 &  0.00    \\
\bottomrule
\end{tabular}
\end{table}

Table~\ref{tab:linear_probe} shows that the \emph{raw} discriminative power of SRM vectors is \textbf{strongly depth-dependent}: Early SRM vectors achieve only 46.17\% accuracy (barely above chance for 47 classes), while Late vectors reach 79.73\%. This 33.56\,pp gap demonstrates that texture-discriminative statistics become progressively more linearly separable with backbone depth.

Combined with the depth-invariance result in Table~\ref{tab:srm_depth}, this reveals a key mechanistic property: \emph{the interact$\rightarrow$distill fusion head can extract equivalent classification performance from SRM vectors of vastly different intrinsic quality}. Even when the SRM signal is weak (46\% linear separability), the cross-order self-attention mechanism compensates by leveraging the complementary spatial token stream.

\subsubsection{Representational Similarity Analysis}
\label{sec:rsa}

To understand the geometric relationship between SRM vectors at different depths, we compute pairwise linear CKA~\cite{Kornblith2019} and mean sample-wise Pearson correlation on the test set.

\begin{table}[t]
\centering
\caption{Pairwise similarity between SRM vectors extracted at different backbone depths on DTD. (a) Linear CKA. (b) Mean sample-wise Pearson correlation.}
\label{tab:rsa}
\setlength{\tabcolsep}{4pt}
\renewcommand{\arraystretch}{1.05}

\subfloat[Linear CKA\label{tab:rsa_cka}]{%
\begin{minipage}{0.48\columnwidth}
\centering
\begin{tabular}{lccc}
\toprule
      & Early & Mid   & Late  \\
\midrule
Early & 1.000 & 0.742 & 0.304 \\
Mid   & 0.742 & 1.000 & 0.492 \\
Late  & 0.304 & 0.492 & 1.000 \\
\bottomrule
\end{tabular}
\end{minipage}
}
\hfill
\subfloat[Mean Pearson correlation\label{tab:rsa_pearson}]{%
\begin{minipage}{0.48\columnwidth}
\centering
\begin{tabular}{lccc}
\toprule
      & Early & Mid   & Late  \\
\midrule
Early & 1.000 & 0.317 & 0.004 \\
Mid   & 0.317 & 1.000 & 0.026 \\
Late  & 0.004 & 0.026 & 1.000 \\
\bottomrule
\end{tabular}
\end{minipage}
}

\end{table}

The results in Table~\ref{tab:rsa} reveal that SRM vectors at different depths are \emph{not redundant}, they occupy geometrically distinct subspaces. Early--Late CKA is only 0.304, and sample-wise Pearson correlation is nearly zero ($r = 0.004$). This means the depth invariance observed in the full model is not a trivial consequence of representational redundancy: the backbone encodes genuinely different second-order statistics at each depth, yet the fusion mechanism integrates them equally well.

Taken together, the linear probe and RSA results paint a coherent picture: modern self-supervised ViTs progressively refine texture-discriminative statistics across depth (Table~\ref{tab:linear_probe}), encoding them in increasingly distinct geometric subspaces (Table~\ref{tab:rsa}), while HiPerViT's interact$\rightarrow$distill topology robustly integrates these statistics regardless of their origin or quality (Table~\ref{tab:srm_depth}).

\subsubsection{Corruption Robustness}
\label{sec:corruption}

Finally, we test whether depth sensitivity emerges under image degradation. Using a model trained with the SRM@Late configuration, we apply test-time corruptions at three severity levels and re-evaluate with the SRM source depth swapped to early, mid, or late.

\begin{table}[t]
\centering
\caption{Top-1 accuracy (\%) under test-time corruptions on DTD, with SRM source depth swapped at inference. Parentheses show degradation from the clean baseline (83.09\%). The checkpoint was trained with SRM@Late.}
\label{tab:corruption}

\footnotesize
\setlength{\tabcolsep}{2.8pt}
\renewcommand{\arraystretch}{1.03}

\begin{tabular}{@{}llccc@{}}
\toprule
\textbf{Corruption} & \textbf{Sev.} &
\textbf{Early} & \textbf{Mid} & \textbf{Late} \\
\midrule

\multirow{3}{*}{Gaussian noise}
& Mild & 83.35 (+0.3) & 83.40 (+0.3) & 83.51 (+0.4) \\
& Med. & 83.03 (-0.1) & 83.30 (+0.2) & 83.19 (+0.1) \\
& Sev. & 82.18 (-0.9) & 82.66 (-0.4) & 82.13 (-1.0) \\

\midrule
\multirow{3}{*}{Gaussian blur}
& Mild & 82.87 (-0.2) & 82.87 (-0.2) & 82.87 (-0.2) \\
& Med. & 82.39 (-0.7) & 82.39 (-0.7) & 82.39 (-0.7) \\
& Sev. & 81.76 (-1.3) & 81.76 (-1.3) & 81.76 (-1.3) \\

\midrule
\multirow{3}{*}{JPEG compr.}
& Mild & 81.44 (-1.6) & 81.44 (-1.6) & 81.44 (-1.6) \\
& Med. & 74.26 (-8.8) & 74.26 (-8.8) & 74.26 (-8.8) \\
& Sev. & 48.19 (-34.9) & 48.19 (-34.9) & 48.19 (-34.9) \\

\midrule
\multirow{3}{*}{Contrast red.}
& Mild & 83.14 (+0.1) & 83.14 (+0.1) & 83.14 (+0.1) \\
& Med. & 82.82 (-0.3) & 82.82 (-0.3) & 82.82 (-0.3) \\
& Sev. & 76.81 (-6.3) & 76.81 (-6.3) & 76.81 (-6.3) \\

\bottomrule
\end{tabular}

\end{table}

Table~\ref{tab:corruption} confirms that \textbf{depth invariance persists under corruption}. For blur, JPEG, and contrast, all three SRM depths yield identical accuracy at every severity. Under Gaussian noise, minor fluctuations appear ($<$0.5\,pp spread), but no systematic advantage for any particular depth emerges. This indicates that the fusion head's ability to compensate for SRM quality is robust even when input quality degrades substantially.

\vspace{0.5em}
\noindent\textbf{Summary.}
These four analyses collectively demonstrate that: (1) raw SRM discriminability increases substantially with backbone depth; (2) SRM vectors at different depths encode genuinely different statistics; yet (3) HiPerViT's interact$\rightarrow$distill fusion achieves invariant final performance across depths, even under corruption. This establishes that the architectural topology, not the source depth of second-order statistics, is the primary mechanism underlying HiPerViT's texture recognition capability.

\subsection{Cue-Conflict Behavior and Conditional Texture Utilization}
\label{sec:cue_conflict}

The mechanistic analyses above demonstrate that HiPerViT's fusion topology robustly integrates second-order statistics regardless of their intrinsic quality.
A complementary question is whether this integration \emph{alters the model's perceptual bias}, specifically, whether injecting an explicit texture channel causes the model to rely more on texture cues at the expense of shape.

We evaluate this using cue-conflict images generated via AdaIN style transfer~\cite{Huang2017} on ImageNet-100: each image preserves the global shape (silhouette) of one class while adopting the texture statistics of a different class.
For each prediction, we determine whether the model matched the shape label $y^{s}$, the texture label $y^{t}$, or neither, and compute three metrics:
\begin{align}
\mathrm{SBI} &= \frac{1}{N}\sum_{i=1}^{N} \mathbf{1}[\hat{y}_i = y_i^{s}], \\
\mathrm{TBI} &= \frac{1}{N}\sum_{i=1}^{N} \mathbf{1}[\hat{y}_i = y_i^{t}], \\
\mathrm{CA}  &= \frac{1}{N}\sum_{i=1}^{N} \mathbf{1}[\hat{y}_i \in \{y_i^{s}, y_i^{t}\}].
\end{align}
where SBI is the Shape-Bias Index, TBI is the Texture-Bias Index, and CA is the overall Conflict Accuracy.

\begin{table}[t]
\centering
\caption{Cue-conflict analysis on ImageNet-100. Both models are trained with identical recipes. $\Delta$TBI is relative to the ViT baseline. Higher TBI indicates stronger texture reliance; higher SBI indicates shape reliance.}
\label{tab:cue_conflict}

\footnotesize
\setlength{\tabcolsep}{3pt}
\renewcommand{\arraystretch}{1.05}

\begin{tabular}{@{}lcccc@{}}
\toprule
\textbf{Model} &
\textbf{SBI (\%)} &
\textbf{TBI (\%)} &
\textbf{CA (\%)} &
$\boldsymbol{\Delta}$\textbf{TBI} \\
\midrule
ViT (baseline) & 88.67 & 1.20 & 89.87 & -- \\
HiPerViT       & 91.47 & 0.53 & 92.00 & $-0.67$ \\
\bottomrule
\end{tabular}

\end{table}

Table~\ref{tab:cue_conflict} reveals a key finding: HiPerViT does \emph{not} increase texture bias relative to a baseline ViT when trained on ImageNet-100.
Both models remain strongly shape-dominant (TBI $< 2$\%), with HiPerViT exhibiting slightly higher shape-bias index and marginally improved overall conflict accuracy.

This result clarifies an important property of the interact$\rightarrow$distill topology:
the statistical token does \emph{not} override task-optimal cues. Rather, cross-order attention dynamically amplifies the feature stream that maximizes discriminative signal under the training objective.
On ImageNet, where shape is the statistically dominant cue, this corresponds to global structure alignment.
Combined with our low-data regime analysis (Sec.~\ref{sec:lowdata}) and texture-benchmark results (Sec.~\ref{sec:ablation_components}), these findings support a \emph{conditional-utilization hypothesis}: HiPerViT augments the representational space with second-order statistics without imposing texture bias when texture is not task-relevant.

Formally, let $T$ denote spatial tokens and $S$ the statistical token. The cross-attention output can be expressed as:
\begin{equation}
A(T,S) = \alpha(T,S)\,T + \beta(T,S)\,S,
\end{equation}
where the learned coefficients $\alpha$ and $\beta$ adapt based on task statistics.
Cue-conflict results suggest that $\alpha$ dominates under object-centric supervision, while texture benchmarks increase the contribution of $\beta$.
This behavior is consistent with the depth-invariance result in Sec.~\ref{sec:mechanistic}, where we show that final accuracy is invariant to SRM intrinsic discriminability.
The interact$\rightarrow$distill topology thus acts as an \emph{adaptive gating mechanism} rather than a static statistical augmentation.

\subsection{Cross-Dataset Transfer and Representation Specialization}
\label{sec:transfer}

The mechanistic analyses in Sec.~\ref{sec:mechanistic} demonstrate that HiPerViT's interact$\rightarrow$distill topology robustly integrates second-order statistics regardless of their intrinsic quality.
A natural follow-up question is whether this fusion mechanism preserves the \emph{generic transferability} of the underlying backbone or induces task-specific alignment that reduces cross-domain generality.

To investigate this, we conduct a controlled cross-dataset transfer study between DTD (47 classes, general textures) and GTOS-Mobile (31 classes, outdoor ground textures).
We train HiPerViT on a source dataset and evaluate transfer via linear probing on the target dataset, comparing representations extracted at four progressively more specialized stages:
\begin{enumerate}
    \item \textbf{Frozen DINOv2 CLS:} the [CLS] token from the original pretrained backbone (no fine-tuning);
    \item \textbf{Frozen DINOv2 Multi-stage:} concatenated GAP features at the same capture blocks used by HiPerViT, from the frozen backbone;
    \item \textbf{HiPerViT Backbone:} multi-stage features from the \emph{fine-tuned} backbone, extracted \emph{before} the SRM and fusion head;
    \item \textbf{HiPerViT Latent:} the post-distillation latent representation after interact$\rightarrow$distill fusion.
\end{enumerate}

\begin{table}[t]
\centering
\caption{Cross-dataset transfer accuracy (\%) via linear probing. Features are extracted at four representation stages of increasing architectural specialization.}
\label{tab:transfer}

\footnotesize
\setlength{\tabcolsep}{4pt}
\renewcommand{\arraystretch}{1.05}

\begin{tabular}{@{}lcccc@{}}
\toprule
\textbf{Direction} &
\textbf{CLS} &
\textbf{Multi} &
\textbf{Pre-Fus.} &
\textbf{Latent} \\
\midrule
DTD $\rightarrow$ GTOS &
\textbf{90.74} &
90.01 &
90.06 &
80.14 \\

GTOS $\rightarrow$ DTD &
\textbf{81.22} &
\textbf{81.22} &
72.71 &
64.10 \\
\bottomrule
\end{tabular}

\end{table}

Table~\ref{tab:transfer} reveals a clear representational gradient.
When source diversity is high (DTD$\rightarrow$GTOS), the fine-tuned backbone features \emph{before} fusion match frozen DINOv2 performance (90.06\% vs.~90.01--90.74\%), indicating that supervised training with hierarchical fusion does not degrade backbone-level universality under favorable source conditions.
In contrast, the post-distillation latent features show reduced cross-domain separability (80.14\%), confirming that the interact$\rightarrow$distill bottleneck induces task-adaptive alignment of second-order correlations.

The reverse direction (GTOS$\rightarrow$DTD) exhibits stronger degradation in both pre-fusion and latent features (72.71\% and 64.10\%), consistent with over-specialization under limited source diversity, GTOS contains fewer classes and a narrower visual vocabulary than DTD.

These results admit a principled interpretation.
Let $\mathbf{f}_b$ denote backbone features and $\mathbf{f}_l$ the latent features after cross-attention distillation.
Because the latent queries $\mathbf{L}$ are optimized under source supervision, the cross-attention operator $A(\mathbf{L}, \mathbf{Z}_\text{vis})$ induces a source-conditioned projection of feature correlations.
Consequently, $\mathbf{f}_l$ becomes a distribution-aligned representation rather than a domain-agnostic embedding, which explains why $\mathbf{f}_b$ retains transferability whereas $\mathbf{f}_l$ exhibits task specialization.

These findings complement the depth-invariance analysis (Sec.~\ref{sec:mechanistic}): while the interact$\rightarrow$distill topology compensates for weak second-order statistics in-domain, it also encourages context-conditioned statistical alignment.
This mechanism improves discriminative separability within the training distribution but reduces isotropic generality across domains.
Together, these results highlight a controllable trade-off between universality and specialization: HiPerViT preserves general backbone representations while enabling explicit statistical alignment through its fusion bottleneck, offering practitioners a choice between generic representation reuse and domain-specific optimization.

\subsection{Computational Efficiency}\label{sec:efficiency}

\begin{figure*}[!ht]
    \centering
    \includegraphics[width=\textwidth]{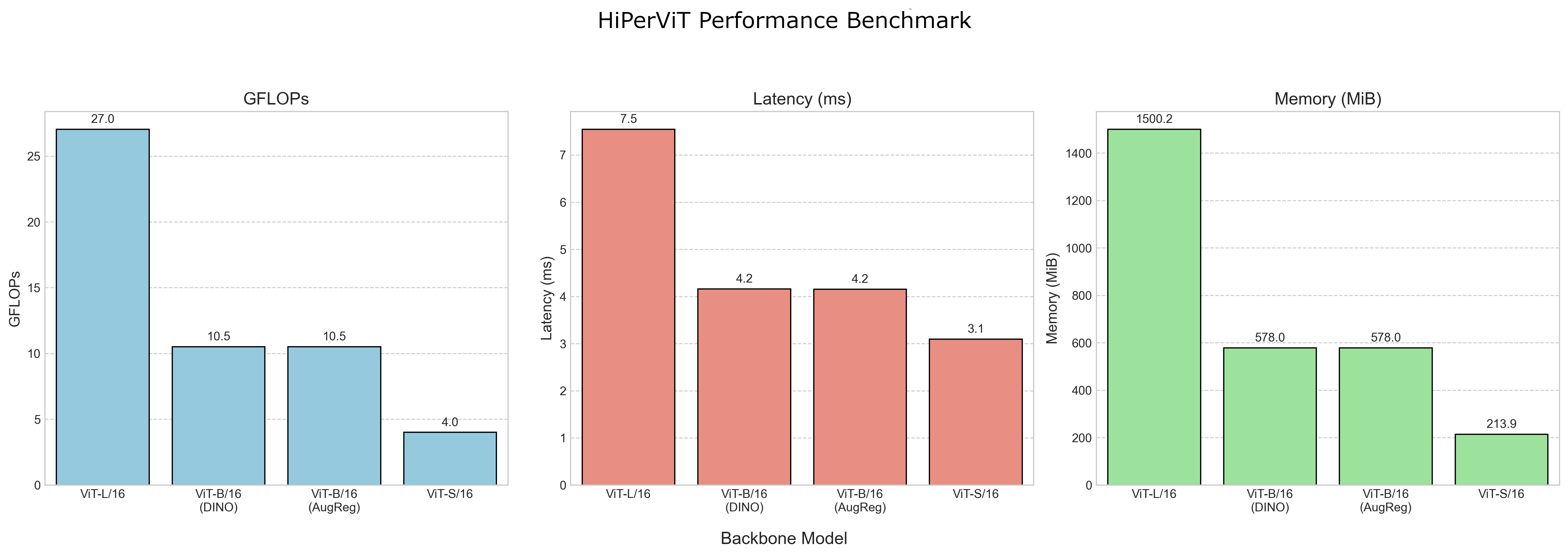}
    \caption{Computational benchmarking of HiPerViT with different ViT backbones. Metrics include GFLOPs, inference latency, and peak GPU memory.}
    \label{fig:benchmark}
\end{figure*}

Finally, we benchmark HiPerViT across four ViT backbones (S, B, L) under the best-performing configuration. Figure \ref{fig:benchmark} reports FLOPs, latency, and peak GPU memory.

The ViT-L/16 variant is most expensive (27.0 GFLOPs), yet accuracy gains over smaller backbones are modest. In contrast, ViT-B/16 reduces FLOPs by over 60\% while maintaining nearly identical accuracy. Remarkably, ViT-S/16 achieves strong performance at only 4.0 GFLOPs and 214 MiB peak memory, making HiPerViT viable for resource-constrained deployment.

Crucially, when compared to prior SOTA, efficiency improvements are stark: ConvNeXt-XXL with RADAM requires $>$350 GFLOPs yet fails to outperform HiPerViT+ViT-B on FMD or GTOS-Mobile. Thus, HiPerViT not only advances accuracy but does so with orders-of-magnitude lower cost.

\begin{figure*}[!ht]
    \centering
    \includegraphics[width=0.8\linewidth]{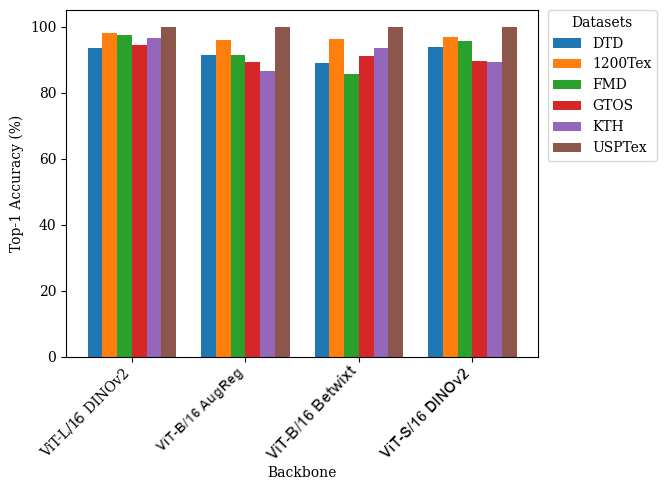}
    \caption{HiPerViT accuracy (\%) on six benchmarks across four ViT backbones. ViT-L/14 DINOv2 achieves the highest mean accuracy, but smaller backbones remain competitive.}
    \label{fig:back_comp}
\end{figure*}

In summary, HiPerViT consistently outperforms vision-only baselines across datasets, with particularly large margins on fine-grained textures. Its design—multi-scale input, cross-stage fusion, and bilinear pooling—proves more effective and efficient than brute-force backbone scaling. The architecture delivers both SOTA accuracy and superior computational efficiency, underscoring its suitability for real-world texture recognition.

\subsection{Real-world applications}
\label{sec:real_world}

While Sections~5.1--5.4 validate HiPerViT on established texture benchmarks, a natural question is whether the proposed extract$\rightarrow$distill$\rightarrow$interact paradigm transfers to \emph{real} scientific settings, where (i) acquisition conditions are domain-specific, (ii) sample sizes may be limited, and (iii) specialized sensors (e.g., hyperspectral/multispectral systems) are not always available.
To probe this, we evaluate HiPerViT on three datasets \emph{introduced in recent peer-reviewed works} spanning biotechnology (plant-based pollution sensing), medicine (prostate histopathology), and agriculture (greenhouse tomato disease/pest monitoring). Importantly, in all three cases HiPerViT is applied \emph{without domain-specific architectural tailoring} and achieves performance that matches or surpasses the results reported in the original publications.

\vspace{0.25em}
\paragraph{\textbf{Biotechnology:}}
We consider hyperspectral confocal microscopy images of \textit{Jacaranda caroba} leaves exposed to different potassium fluoride pollutant levels, recently investigated in \cite{DNAS_CBM}.
That work explicitly targets \emph{whole-image} classification of high-resolution hyperspectral data under limited sample size and high dimensionality, proposing a hand-engineered complex-network descriptor (DNAS) with 32 features and reporting \textbf{92.6\%} accuracy.
In contrast, HiPerViT operates on \emph{standard RGB} inputs, i.e., it does not assume access to hyperspectral/multispectral instrumentation. To ensure a strictly fair methodological comparison, we evaluate HiPerViT using the exact same image cross-validation folds/splits as the DNAS paper, but process only the 3 corresponding visible RGB bands rather than the full 32-channel spectral cube.
Despite this substantially weaker sensing modality, HiPerViT reaches \textbf{95.0\%} accuracy on the identical classification task.
This result is practically significant: it suggests that the representational bias induced by hierarchical multi-scale extraction and explicit second-order interactions can compensate, at least in part, for the absence of rich spectral bands. We hypothesize that by encoding pairwise feature correlations (via the SRM), the model learns to exploit subtle color-spatial dependencies that act as ``pseudo-spectral'' signatures, effectively recovering discriminative information that would otherwise require hyperspectral instrumentation. This enables more accessible deployments in resource-limited monitoring scenarios.

\vspace{0.25em}
\paragraph{\textbf{Medicine:}}
Next, we evaluate on SICAPv2 prostate biopsy images, a challenging histology benchmark used in \cite{Gleason_End2End_Elsevier} and also adopted in \cite{Gleason_Arxiv_Comparative}.
The clinical motivation is strong: Gleason grading and the identification of grade-4 patterns (including cribriform morphology) are time-consuming and subject to inter-observer variability, motivating robust automated decision support. Reported results on SICAPv2 include \textbf{76.22\%} accuracy in \cite{Gleason_End2End_Elsevier} and \textbf{85.13\%} accuracy in \cite{Gleason_Arxiv_Comparative}. To guarantee no data leakage and maintain parity, we strictly follow the official patient-level train/test cross-validation splits provided by the dataset authors. Furthermore, we operate directly on the authors' pre-extracted histological tiles (patches) and apply standard ImageNet RGB normalization without any domain-specific stain normalization routines (e.g., Macenko).
Under this rigorous evaluation setup, HiPerViT achieves \textbf{87.1\%} accuracy, exceeding both reported baselines.
Beyond the numeric improvement, this experiment highlights that HiPerViT can capture diagnostically relevant micro-architectural patterns in H\&E tissue, a domain where discriminative cues often manifest as subtle textural organization across multiple scales.

\vspace{0.25em}
\paragraph{\textbf{Agriculture:}}
Finally, we test HiPerViT on the tomato leaf disease dataset introduced in \cite{TLID_JBCS}, which reflects region-specific disease occurrence under greenhouse cultivation.
That study reports \textbf{90.48\%} accuracy when operating on a patch-based variant (PTLID), and \textbf{86.56\%} accuracy on the original non-patch images (TLID).
Here we focus on the \emph{non-patch} setting to assess performance without the additional supervision and engineering implicit in patch extraction. Crucially, we utilize the exact same training and testing data splits as the original paper to ensure a direct comparison.
In this configuration, HiPerViT attains \textbf{87.5\%} accuracy, surpassing the original non-patch baseline.
This is notable because greenhouse imagery often exhibits complex nuisance factors (illumination variability, cluttered backgrounds, heterogeneous symptom stages), making generalization beyond curated datasets particularly challenging.

\vspace{0.25em}
\paragraph{\textbf{Summary and implications.}}
Table~\ref{tab:real_world_apps} summarizes these results.
Across three distinct real-world domains, HiPerViT consistently improves upon the best reported numbers from the corresponding sources, including scenarios where the published baselines rely on specialized sensing (hyperspectral/multispectral) or task-specific preprocessing (patch pipelines).
Taken together, these findings support a central claim of this work: explicitly structuring computation into (i) multi-scale feature preservation, (ii) attention-based token distillation, and (iii) second-order interaction modeling yields a representation that is not only competitive on texture benchmarks, but also \emph{deployable and robust} in practical scientific applications where data are limited and acquisition constraints are non-ideal.

\begin{table}[t]
\centering
\footnotesize
\caption{Real-world applications. HiPerViT compared with results reported in the original publications for three datasets spanning biotechnology, medicine, and agriculture.}
\label{tab:real_world_apps}
\setlength{\tabcolsep}{3pt}
\renewcommand{\arraystretch}{1.08}
\begin{tabularx}{\columnwidth}{@{}>{\raggedright\arraybackslash}p{0.31\columnwidth} >{\centering\arraybackslash}p{0.15\columnwidth} >{\centering\arraybackslash}p{0.13\columnwidth} >{\raggedright\arraybackslash}X@{}}
\toprule
\textbf{Domain / Dataset} & \textbf{Best} & \textbf{HiPerViT} & \textbf{Notes} \\
\midrule
Biotech / \textit{J.\ caroba} pollution
& 92.6\%
& 95.0\%
& Baseline uses hyperspectral/multispectral data; HiPerViT uses RGB. \\

Medicine / SICAPv2
& 85.13\% / 76.22\%
& 87.1\%
& Histopathology grading-related classification. \\

Agriculture / TLID
& 86.56\%
& 87.5\%
& Uses non-patch TLID images, not PTLID patches. \\
\bottomrule
\end{tabularx}

\vspace{0.2em}
\begin{minipage}{\columnwidth}
\footnotesize
\textit{Sources:} \textit{J.\ caroba} pollution \cite{DNAS_CBM}; SICAPv2 \cite{Gleason_End2End_Elsevier,Gleason_Arxiv_Comparative}; TLID \cite{TLID_JBCS}.
\end{minipage}
\end{table}

\paragraph{\textbf{Summary and implications}:}
Table~\ref{tab:real_world_apps} consolidates the outcomes across biotechnology, medicine, and agriculture.
Despite the pronounced domain shift, from confocal hyperspectral microscopy to H\&E whole-slide histology and greenhouse leaf imagery, HiPerViT remains consistently competitive and, in these cases, exceeds the best numbers reported in the corresponding sources.
Two aspects are particularly noteworthy.
First, HiPerViT improves over pipelines that rely on stronger sensing modalities (hyperspectral/multispectral) or additional task-specific engineering (e.g., patch-based supervision), indicating that the proposed representation is not merely exploiting dataset idiosyncrasies but capturing transferable visual structure.
Second, the gains are achieved without domain-specific architectural modifications, supporting the claim that the inductive bias introduced by our hierarchical multi-scale extraction, attention-based token distillation, and explicit second-order interactions is broadly applicable.

\vspace{0.25em}
\paragraph{\textbf{Closing remark:}}
Taken together, these experiments strengthen the central message of this paper: \emph{second-order statistical structure is a foundational visual primitive}, and explicitly reintroducing it into SSL transformer representations via Statistical Token Injection yields gains that generalize beyond canonical texture benchmarks.
Crucially, this design offers a practical accuracy--efficiency trade-off: by distilling informative multi-scale tokens and encoding their structured interactions, HiPerViT delivers robust performance in heterogeneous, real-world scientific pipelines while operating under realistic constraints on data volume and sensing hardware.

\section{Discussion}
HiPerViT establishes consistent improvements over prior state-of-the-art models across five of six benchmarks (Table~\ref{tab:sota_comparison}), demonstrating the practical and conceptual value of Statistical Token Injection within a multi-scale extraction and latent distillation framework. Its multi-scale token hierarchy preserves both local micro-patterns and global context, while the compact bilinear head efficiently encodes pairwise feature correlations. The Perceiver-style latent bottleneck integrates information asymmetrically, condensing tens of thousands of tokens into a manageable set of latent queries. Collectively, these mechanisms, unified through the STI principle, allow HiPerViT to deliver robust recognition without relying on excessively large backbone architectures.

The gains are particularly pronounced on GTOS-Mobile (+10.48 pp) and 1200Tex (+10.27 pp), highlighting that the model excels on high-resolution, domain-specific textures where second-order statistics carry strong discriminative signal. On more challenging and unconstrained datasets, such as DTD, HiPerViT maintains a strong +3.05 pp improvement over a ViT-L/14 baseline, demonstrating that efficiency and generalization need not be mutually exclusive. Across all datasets, smaller ViT-B/S backbones paired with the HiPerViT head achieve competitive accuracy at substantially reduced FLOPs and memory usage, confirming that our approach balances state-of-the-art performance with practical deployment constraints.

This cross-dataset success suggests that injecting explicit second-order statistical structure into SSL transformer representations constitutes a broadly applicable strategy. Rather than favoring one architectural family, HiPerViT selectively leverages each paradigm where it is strongest: early-stage token hierarchies for local patterns, transformer bottlenecks for global integration, and compact bilinear statistics for feature interactions. This combination yields a resilient architecture that generalizes across diverse visual domains, acquisition conditions, and resolutions.

\subsection{Dual Invariance as Architectural Robustness}

HiPerViT exhibits two complementary invariances that jointly characterize the robustness of its fusion mechanism.
First, depth-invariance (Sec.~\ref{sec:mechanistic}): the fusion head extracts equivalent classification performance from SRM vectors of vastly different intrinsic quality, indicating that cross-order self-attention compensates for weak second-order signals by leveraging the complementary spatial token stream.
Second, topology-invariance (Sec.~\ref{sec:topology_robustness}): interact$\rightarrow$distill, distill$\rightarrow$interact, and late fusion all achieve statistically indistinguishable accuracy on DTD, establishing that the discriminative gain originates from the SRM statistical prior rather than from a brittle wiring choice.
Together, these properties make HiPerViT robust to hyperparameter choices about both \emph{where} to tap second-order statistics from the backbone and \emph{how} to integrate them into the fusion pipeline.

\subsection{Limitations and Future Work}
Despite these advances, the current study exposes clear directions for further research:
\begin{itemize}
    \item \textbf{Counterfactual controls:} Our ablations establish the importance of the SRM statistical token, but do not yet include counterfactual baselines (e.g., random tokens, shuffled SRM, or learned MLP tokens of equivalent capacity). Adding these controls would provide stronger causal evidence for STI.
    \item \textbf{Non-texture domains:} While HiPerViT's gains are validated on texture benchmarks and three applied scientific tasks, broader evaluation on histopathology, remote sensing, and fine-grained biological classification would strengthen the claim of domain generality.
    \item \textbf{Temporal and dynamic textures:} The current evaluation is limited to static images. Extending the STI framework to spatio-temporal domains could enable recognition of dynamic textures, where temporal consistency is critical.
    \item \textbf{Statistical rigor:} Key invariance claims are based on 3 seeds with limited partitions. Expanding to $\geq$5 seeds with 95\% confidence intervals would increase defensibility.
    \item \textbf{Interaction regimes:} Topology ablations show invariance on DTD; identifying regimes where cross-order attention outperforms late fusion (e.g., under heavy domain shift or extreme data scarcity) would strengthen the conditional-advantage narrative.
\end{itemize}

\subsection{When Does STI Help?}

Our experiments suggest that STI is most beneficial when three conditions are jointly met:
(i)~the task's discriminative signal resides in pairwise feature correlations rather than first-order averages;
(ii)~the backbone's pre-training objective does not explicitly encode second-order structure (as is the case for standard SSL-ViTs); and
(iii)~supervision is limited, amplifying the value of a structured inductive bias (Sec.~\ref{sec:lowdata}).
Conversely, STI does not distort representations when these conditions are absent: cue-conflict evaluation on ImageNet-100 (Sec.~\ref{sec:cue_conflict}) confirms that the statistical token's influence is conditioned on task relevance.
We emphasize this as a \emph{conditional} inductive bias: STI enriches the representation without overriding the backbone's dominant priors.

Overall, HiPerViT demonstrates that carefully combining hierarchical, statistical, and attention-based mechanisms can achieve state-of-the-art performance while remaining computationally efficient, providing a strong blueprint for extending SSL transformer representations with explicit statistical structure. 



\section{Conclusion}
We introduced \textbf{Statistical Token Injection (STI)}, a lightweight mechanism for reintroducing second-order statistical structure into self-supervised Vision Transformer representations. STI embeds compact bilinear descriptors as first-class tokens within the transformer's self-attention, enabling cross-order interaction between spatial and statistical features at negligible parameter cost. We validated STI through \textbf{HiPerViT}, a modular architecture pairing STI with multi-scale extraction and Perceiver-style latent distillation. Evaluated on six texture benchmarks and three applied scientific domains, HiPerViT achieves state-of-the-art or competitive vision-only performance across the evaluated benchmarks, with gains of up to +10.48\,pp.

Our analysis highlights four key insights. First, transformer scale alone does not substantially drive performance within the evaluated backbone family: the accuracy gap across ViT-S/B/L backbones remains under 2\,pp once STI and the Perceiver head are integrated. Second, HiPerViT achieves a favorable accuracy--efficiency trade-off, maintaining SOTA-level performance while substantially reducing FLOPs relative to higher-capacity backbone configurations. Third, mechanistic analysis shows that final recognition performance is remarkably insensitive to the backbone depth from which the second-order statistics are extracted. Fourth, controlled topology ablations indicate that the benefit is associated more strongly with the availability of the second-order representation than with a particular interaction--distillation ordering. Together, these results support the view that making second-order statistics explicitly available to the representation is more important than the particular wiring used to integrate them.

These findings suggest a broader principle for the machine intelligence community: explicit statistical inductive biases can be incorporated into SSL transformer representations through lightweight integration mechanisms, providing consistent benefits in tasks where micro-structural information is discriminative, without requiring domain-specific architectures, multimodal pre-training, or excessive compute.

$\,$

$\,$

\bibliographystyle{plain}
\bibliography{references}
\end{document}